\documentclass[letterpaper,10pt,conference]{ieeeconf}
\IEEEoverridecommandlockouts
\useRomanappendicesfalse

\usepackage{amsmath,amssymb,mathtools}
\usepackage{booktabs}
\usepackage{multirow}
\usepackage{array}
\usepackage{graphicx}
\usepackage{xcolor}
\usepackage{microtype}
\usepackage{cite}
\usepackage{url}
\usepackage{hyperref}
\usepackage{placeins}
\usepackage{balance}
\usepackage{eso-pic}
\usepackage{algpseudocode}
\usepackage{tikz}
\usetikzlibrary{arrows.meta,calc,fit,positioning,shapes.geometric}

\graphicspath{{figures/}}
\hypersetup{hidelinks,
  pdftitle={HA-DiPO: Diffusion Policy Optimization for Heterogeneous Cooperative Transport},
  pdfauthor={Xuebin Li, Jinpeng Xiao, Zong Chen, Pierluigi Arpenti, Bruno Siciliano, Jiebao Sun, and Yiqun Li},
  pdfsubject={Preprint submitted to IEEE Robotics and Automation Letters},
  pdfkeywords={cooperative transport, heterogeneous robots, diffusion policies, multi-agent reinforcement learning}}

\definecolor{hadipoblue}{RGB}{40,95,145}
\definecolor{hadipogreen}{RGB}{45,125,95}
\definecolor{hadipoorange}{RGB}{205,120,45}
\definecolor{draftred}{RGB}{170,25,25}

\newif\ifdraftnotes
\draftnotestrue

\newcommand{\method}{HA-DiPO}
\newcommand{\proxyQ}{\overline{Q}}
\newcommand{\proxyV}{\overline{V}}
\newcommand{\proxyJ}{\overline{J}}
\newcounter{algorithm}
\algrenewcommand{\algorithmicrequire}{\textbf{Input:}}

\title{\LARGE\bfseries \method: Diffusion Policy Optimization for Heterogeneous \\ Cooperative Transport}

\author{Xuebin~Li$^{1}$, Jinpeng~Xiao$^{2}$, Zong~Chen$^{2}$, Pierluigi~Arpenti$^{3}$,\\
Bruno~Siciliano$^{3}$, Jiebao~Sun$^{1}$, and Yiqun~Li$^{2,*}$
\thanks{{$^{1}$Xuebin Li and Jiebao Sun are with the School of Mathematics, Harbin Institute of Technology, Harbin, China (e-mail: \nolinkurl{23B912027@stu.hit.edu.cn}; \nolinkurl{sunjiebao@hit.edu.cn}).}}
\thanks{{$^{2}$Jinpeng Xiao, Zong Chen, and Yiqun Li are with the State Key Laboratory of Intelligent Manufacturing Equipment and Technology, School of Mechanical Science and Engineering, Huazhong University of Science and Technology, Wuhan, China (e-mail: \nolinkurl{m202570609@hust.edu.cn}; \nolinkurl{skelon_chan@hust.edu.cn}; \nolinkurl{liyiqun_hit@163.com}).}}
\thanks{{$^{3}$Pierluigi Arpenti and Bruno Siciliano are with the Department of Electrical Engineering and Information Technology, University of Naples Federico II, Naples, Italy (e-mail: \nolinkurl{pierluigi.arpenti@unina.it}; \nolinkurl{siciliano@ieee.org}).}}
\thanks{{$^{*}$Corresponding author: Yiqun Li (\nolinkurl{liyiqun_hit@163.com}). Project website and demonstration videos: \url{https://ha-dipo.github.io/}.}}
}

\begin{document}
% First-page preprint notice, below the manuscript text area.
\AddToShipoutPictureFG*{\AtPageLowerLeft{\put(0,0){\makebox[\paperwidth][c]{\raisebox{25pt}{\parbox[b]{\textwidth}{\centering\footnotesize This work has been submitted to IEEE Robotics and Automation Letters (RA-L).}}}}}}
\bstctlcite{ral-bib-control}
\maketitle
\thispagestyle{empty}
\pagestyle{empty}

\begin{abstract}
Cooperative transport with heterogeneous legged robots requires coordinating different motion capabilities while supporting an unfixed shared payload.
We present \method{} (Heterogeneous-Agent Diffusion Policy Optimization), a hierarchical multi-agent reinforcement learning framework under centralized training with decentralized execution (CTDE).
The framework decouples team-level coordination from embodiment-specific motion control: at the high level, a centralized critic uses team-value estimates to jointly train independent coordination policies, aligning robot-specific action responses with a shared transport objective; at the low level, embodiment-specific robust motion policies track the coordination commands.
The high-level diffusion policies offer more flexible action distributions than single-Gaussian parameterizations, allowing potentially multimodal coordination responses to be represented.
The coordination policies learn from scratch through reward-driven simulator interaction without demonstrations.
At execution, each robot generates coordination commands using only local robot and payload observations, without partner-state inputs or inter-robot messages.
Simulation comparisons show that \method{} achieves higher final mean returns and velocity-tracking success rates, as well as lower velocity errors, than the evaluated baselines.
Real-world experiments further demonstrate quadrupedal--wheeled-legged and human--robot cooperative transport.
\end{abstract}

\section{Introduction}
\label{sec:introduction}

Cooperative transport of long or heavy objects is an important application of multi-robot systems in industrial and field operations~\cite{tuci2018review}.
Robots with different motion capabilities can share the load, but must coordinate their motion while maintaining stable support.
When the payload rests on the robots without rigid attachment, their actions are coupled through contact: acceleration, slip, or a posture adjustment by one robot can alter the payload state and its partners' support conditions.
Differences in actuation and motion response further complicate coordination, making identical commands insufficient to ensure consistent payload motion.
We study how such robots can learn to coordinate through interaction and transport an unfixed payload using local observations.

However, extending existing transport approaches to heterogeneous teams carrying an unfixed payload remains challenging.
First, some transport systems rely on predefined formation geometry or dedicated coupling mechanisms~\cite{koung2021hqp,turrisi2024pacc}.
These approaches embed coordination in particular geometric constraints or mechanical interfaces, which must be reconsidered when support contacts can shift.
Other systems rely on external motion capture for state feedback~\cite{zhang2026cognition}.
Second, continuous-action multi-agent learning commonly uses single-Gaussian policy parameterizations, which offer limited flexibility for representing multimodal coordination behaviors~\cite{zhang2026ddpl}.
Diffusion policies offer richer action-distribution representations~\cite{chi2025diffusion}, but demonstration-driven diffusion approaches to cooperative transport remain dependent on the availability and coverage of expert data~\cite{li2026hclm}.
Finally, decentralized methods based on parameter sharing or single-robot policy reuse~\cite{pandit2026decplm,cao2026paint} do not explicitly address joint learning of robot-specific policies that accommodate different motion capabilities while coordinating through the shared payload.

\begin{figure*}[t]
  \centering
  \resizebox{\textwidth}{!}{%
\begin{tikzpicture}[
 x=1mm,y=.78mm,font=\footnotesize,
 >={Latex[length=1.35mm,width=0.9mm]},
 panel/.style={draw=black!18,rounded corners=1.2pt,fill=white,line width=.4pt},
 ptitle/.style={font=\small\bfseries,anchor=west,text=black!84},
 psub/.style={font=\scriptsize,anchor=west,text=black!58},
 blk/.style={draw=black!32,rounded corners=1.1pt,fill=white,align=center,inner sep=1.5pt,font=\scriptsize},
 blkb/.style={blk,draw=hadipoblue!70,fill=hadipoblue!5},
 blkg/.style={blk,draw=hadipogreen!70,fill=hadipogreen!5},
 blko/.style={blk,draw=hadipoorange!70,fill=hadipoorange!5},
 flow/.style={->,draw=black!60,line width=.45pt},
 cond/.style={->,draw=hadipoblue!85,line width=.5pt},
 tflow/.style={->,draw=hadipoorange!90!black,line width=.55pt,dash pattern=on 1.2pt off .8pt}
]

\tikzset{
  hist/.pic={\foreach \dx/\dy/\tint in {0/0/3,.65/.65/6,1.3/1.3/11}{
    \draw[hadipoblue!55,fill=hadipoblue!\tint,line width=.3pt,rounded corners=.3pt]
      (\dx-2.7,\dy-2) rectangle (\dx+1.6,\dy+2);}},
  gamepad/.pic={
    \begin{scope}[x=1mm,y=1mm,
      draw=hadipogreen!90!black,line width=.45pt,
      line cap=round,line join=round]
      \draw[fill=hadipogreen!3]
        (-1.7,1.6)--(1.7,1.6)
        .. controls (2.35,1.65) and (2.75,1.15) .. (2.95,.4)
        --(3.3,-1.15)
        .. controls (3.5,-2.1) and (2.65,-2.4) .. (2.1,-1.8)
        --(1.4,-1.12)
        .. controls (1.25,-.99) and (1.1,-1) .. (.9,-1)
        --(-.9,-1)
        .. controls (-1.1,-1) and (-1.25,-.99) .. (-1.4,-1.12)
        --(-2.1,-1.8)
        .. controls (-2.65,-2.4) and (-3.5,-2.1) .. (-3.3,-1.15)
        --(-2.95,.4)
        .. controls (-2.75,1.15) and (-2.35,1.65) .. (-1.7,1.6)--cycle;
      \begin{scope}[shift={(-1.85,.65)}]
        \fill[hadipogreen!80!black]
          (-.5,-.16)--(-.16,-.16)--(-.16,-.5)--(.16,-.5)
          --(.16,-.16)--(.5,-.16)--(.5,.16)--(.16,.16)
          --(.16,.5)--(-.16,.5)--(-.16,.16)--(-.5,.16)--cycle;
      \end{scope}
      \foreach \dx/\dy in {0/.38,.38/0,0/-.38,-.38/0}{
        \fill[hadipogreen!80!black] (1.85+\dx,.65+\dy) circle[radius=.17];}
      \foreach \x in {-.73,.73}{
        \draw[fill=white,line width=.3pt] (\x,-.33) circle[radius=.5];
        \fill[hadipogreen!90!black] (\x,-.33) circle[radius=.25];}
    \end{scope}},
  network/.pic={
    \foreach \a in {-1.8,0,1.8}{\foreach \b in {-1.8,0,1.8}{
      \draw[hadipoblue!35,line width=.22pt] (-2.6,\a)--(0,\b);
      \draw[hadipoblue!35,line width=.22pt] (0,\a)--(2.6,\b);}}
    \foreach \x in {-2.6,0,2.6}{\foreach \y in {-1.8,0,1.8}{
      \filldraw[fill=white,draw=hadipoblue!85,line width=.33pt] (\x,\y) circle[radius=.48];}}},
  data/.pic={
    \draw[hadipoorange!80,fill=hadipoorange!4,line width=.35pt] (-2.7,-2.4)--(-2.7,2.4) arc[start angle=180,end angle=360,x radius=2.7,y radius=.9] --(2.7,-2.4) arc[start angle=0,end angle=-180,x radius=2.7,y radius=.9];
    \draw[hadipoorange!80,fill=hadipoorange!7,line width=.35pt] (0,2.4) ellipse[x radius=2.7,y radius=.9];
  }
}
\path[use as bounding box] (0,0) rectangle (178,81.218);
\begin{scope}[y=.67mm,yshift=-4.99mm]
  % The central scene illustrates two heterogeneous agents and a shared payload.
  \node[panel,fit={(0,67) (178,102)},inner sep=0pt] {};
  \node[ptitle] at (2,98.4) {(a)~~Heterogeneous cooperative transport};
  \node[anchor=south,inner sep=0pt] at (89,72.3)
    {\includegraphics[height=18.6mm]{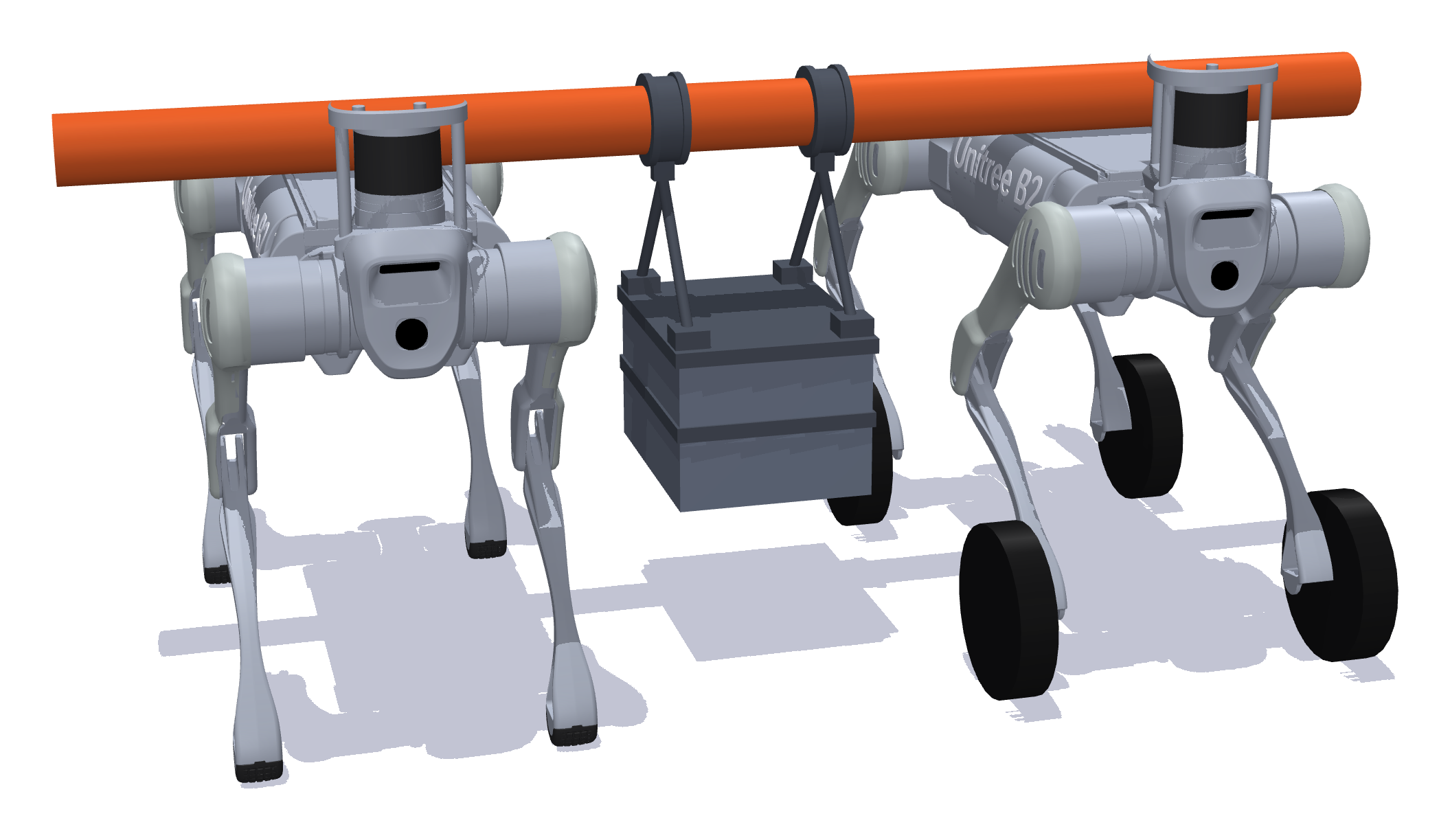}};

  % Task inputs and local execution, with no duplicate control chains.
  \pic at (6.5,89.5) {gamepad};
  \node[blkg,minimum width=46mm,minimum height=5.5mm,font=\footnotesize]
    at (35.5,89.5) {Team command $\mathbf c=[v_x^{\mathrm{ref}},v_y^{\mathrm{ref}},\dot\psi^{\mathrm{ref}}]$};
  \pic at (7,79.8) {hist};
  \node[anchor=west,align=left,font=\footnotesize,text=black!60]
    at (13.8,80.3) {Decentralized execution\\from local observations};

  % Execution constraints and the physical support interface.
  \node[anchor=west,font=\scriptsize,text=black!55]
    at (132,95.4) {At execution};
  \draw[<->,draw=black!40,line width=.4pt] (123,88.7)--(129,88.7);
  \draw[hadipoorange!90!black,line width=.75pt]
    (124.4,87.3)--(127.6,90.1) (124.4,90.1)--(127.6,87.3);
  \node[anchor=west,align=left,font=\footnotesize,text=hadipoorange!90!black]
    at (132,88.7) {\textbf{No inter-agent communication}\\No partner-state inputs};
  \draw[hadipoorange!90!black,line width=.75pt,line cap=round]
    (123,80.7)--(129,80.7);
  \draw[hadipoorange!85!black,fill=hadipoorange!12,line width=.35pt]
    (123.7,78.6)--(124.9,80.5)--(126.1,78.6)--cycle
    (125.9,78.6)--(127.1,80.5)--(128.3,78.6)--cycle;
  \node[anchor=west,align=left,font=\footnotesize,text=hadipoorange!90!black]
    at (132,79.6) {Loose support contact\\No rigid grasp};

  % Compact instance labels identify both policies and the carried load.
  \draw[black!34,line width=.3pt] (65,72.2)--(77.5,80);
  \draw[black!34,line width=.3pt] (89,72.2)--(89,83.5);
  \draw[black!34,line width=.3pt] (114,72.2)--(104,80);
  \node[font=\footnotesize,text=hadipoblue!90!black]
    at (64,69.7) {agent 1 ($\pi_{\theta_1}$)};
  \node[font=\footnotesize,text=hadipoorange!90!black]
    at (89,69.7) {payload};
  \node[font=\footnotesize,text=hadipoblue!90!black]
    at (114,69.7) {agent 2 ($\pi_{\theta_2}$)};
\end{scope}

\node[panel,fit={(0,0)(43,48)},inner sep=0pt,draw=hadipoorange!45,dashed] {};
\node[ptitle] at (2,44.5) {(b)~~Centralized training};
\node[minimum width=35mm,minimum height=7mm] (data) at (21.5,37) {};
\pic at (8,37) {data};
\node[font=\scriptsize,align=center] at (27,37) {Joint replay\\$\mathcal D$};
\node[blko,minimum width=35mm,minimum height=8mm] at (22.3,26.8) {};
\node[blko,minimum width=35mm,minimum height=8mm] (critic) at (21.5,26) {Centralized twin critics\\$Q_\phi^{\min}(\mathbf o,\mathbf a)$};
\draw[tflow] (data)--(critic);
\node[blko,minimum width=35mm,minimum height=4.5mm] (objective) at (21.5,16) {Diffusion-path objective};
\draw[tflow] (critic)--(objective);
\node[blkb,minimum width=9mm,minimum height=5mm] (seq1) at (10,7.8) {$\theta_{i_1}$};
\node[blkb,minimum width=9mm,minimum height=5mm] (seq2) at (24,7.8) {$\theta_{i_2}$};
\node[font=\scriptsize] (seq3) at (37,7.8) {$\cdots$};
\draw[tflow] (objective.south)--(21.5,12)--(10,12)--(seq1.north);
\draw[tflow] (seq1)--(seq2);
\draw[tflow] (seq2)--(seq3);
\node[font=\scriptsize,text=hadipoorange!90!black] at (21.5,2) {Sequential actor updates};
\node[panel,fit={(46,27)(178,48)},inner sep=0pt,draw=hadipoblue!40] {};
\node[ptitle] at (48,44.5) {(c)~~Hierarchical decentralized execution};
\node[anchor=east,font=\scriptsize,text=black!60] at (175,44.5) {Independent parameters $\theta_i$};
\pic at (53,38.3) {hist};
\node[align=center,font=\scriptsize,text=black!65] at (66,38.8) {Observation\\history};
\pic at (53,30.7) {gamepad};
\node[align=center,font=\scriptsize] at (66,30.7)
 {{\color{hadipogreen!90!black}Team command}\\$\mathbf c$};
\draw[flow] (74,38.8)--(76,38.8)--(76,34.5)--(82,34.5);
\node[font=\scriptsize,text=hadipoblue!90!black] at (79,36.6) {$o_i$};
\draw[black!60,line width=.45pt] (74,30.7)--(76,30.7)--(76,34.5);
\node[blkb,minimum width=28mm,minimum height=9mm] (actor) at (96,34.5)
  {\textbf{5 Hz}~$\cdot$~Diffusion actor\\$\pi_{\theta_i}$};
\node[blkg,minimum width=35mm,minimum height=9mm] (tracker) at (137.5,34.5)
  {\textbf{50 Hz}~$\cdot$~Robust policy\\Embodiment-specific\\PPO-pretrained; frozen};
\node[blk,minimum width=17mm,minimum height=9mm] (out) at (167.5,34.5)
  {\textbf{200 Hz}\\Joint-level\\control};
\draw[flow] (actor)--node[above,font=\scriptsize] {$a_i$}(tracker);
\draw[flow] (tracker)--(out);
\draw[tflow] (seq3.east)--(44.5,7.8)--(44.5,25.5)--(96,25.5)--(actor.south);
% A dashed leader identifies (d) as the internal view of the same actor in (c).
\draw[hadipoblue!70,line width=.45pt,dash pattern=on 1.2pt off .9pt]
 (actor.south east)--(actor.south east |- 0,26)--(118,26)--(118,24);
\node[panel,fit={(46,0)(178,24)},inner sep=0pt,draw=hadipoblue!50] {};
\node[ptitle] at (48,20.8) {(d)~~Diffusion actor};
\node[font=\scriptsize,text=hadipoblue!90!black] (obs) at (94,20.8) {$o_i$};
\pic at (110,20.8) {network};
\draw[cond] (obs)--(106,20.8);
\node[anchor=west,font=\scriptsize,text=hadipoblue!90!black] at (115,20.8) {$f_{\theta_i}$};
\draw[hadipoblue!75,line width=.42pt] (110,18.3)--(110,15.2)--(58.5,15.2);
\foreach \x in {58.5,67.5,76.5}{\draw[cond] (\x,15.2)--(\x,9.1);}
\node[font=\scriptsize,text=black!65] at (54,15.5) {Noise};
\node[blk,draw=black!20,minimum width=8mm,minimum height=6.8mm,inner sep=0pt] (stage0) at (54,9.1) {};
\begin{scope}[shift={(54,9.1)},scale=.9]
\foreach \r/\op in {3/.08,2.2/.14,1.4/.23,.65/.34}{\fill[hadipoblue,opacity=\op] (0,0) circle[radius=\r];}
\end{scope}
\node[blk,draw=black!20,minimum width=8mm,minimum height=6.8mm,inner sep=0pt] (stage1) at (63,9.1) {};
\begin{scope}[shift={(63,9.1)},scale=.9]
\foreach \r/\op in {2.8/.09,1.9/.15,1.15/.24,.5/.38}{\fill[hadipoblue,opacity=\op] (0,0) ellipse[x radius=\r*1.25,y radius=\r*.73];}
\end{scope}
\node[blk,draw=black!20,minimum width=8mm,minimum height=6.8mm,inner sep=0pt] (stage2) at (72,9.1) {};
\begin{scope}[shift={(72,9.1)},scale=.9]
\foreach \r/\op in {2/.12,1.25/.22,.55/.36}{\fill[hadipoblue,opacity=\op] (-1.1,.65) circle[radius=\r];\fill[hadipoblue,opacity=\op] (1.1,-.65) circle[radius=\r];}
\end{scope}
\node[blk,draw=black!20,minimum width=8mm,minimum height=6.8mm,inner sep=0pt] (stage3) at (81,9.1) {};
\begin{scope}[shift={(81,9.1)},scale=.9]
\foreach \r/\op in {1.8/.14,1.05/.28,.45/.48}{\fill[hadipoblue,opacity=\op] (-2,.85) circle[radius=\r];\fill[hadipoblue,opacity=\op] (2,-.85) circle[radius=\r];}
\end{scope}
\draw[flow] (stage0.east)--(stage1.west);
\draw[flow] (stage1.east)--(stage2.west);
\draw[flow] (stage2.east)--(stage3.west);
\node[font=\scriptsize,text=black!65] at (54,2.4) {$z_i^N$};
\node[font=\scriptsize,text=black!65] at (81,2.4) {$z_i^0$};

\node[blkb,minimum width=7mm,minimum height=5.5mm] (bound) at (90,9.1) {$T_i$};
\node[font=\scriptsize,text=black!60] at (90,2.4) {tanh + scale};
\draw[flow] (stage3)--(bound);
\node[font=\scriptsize,inner sep=.5pt] (action) at (113,9.1) {$a_i=[v_{x,i},v_{y,i},\dot\psi_i,h_{b,i},\theta_{b,i}]$};
\draw[flow] (bound)--(action);
\draw[black!13,line width=.3pt] (131.5,2)--(131.5,22);

% Schematic responses: illustrative possibilities, not empirical densities.
\node[draw=hadipoorange!35,fill=hadipoorange!3,rounded corners=1pt,line width=.35pt,
 minimum width=20.4mm,minimum height=18.1mm,inner sep=0pt] at (144,12) {};
\node[font=\scriptsize\bfseries,text=hadipoorange!90!black] at (144,20.6) {Gaussian};
\draw[black!30,line width=.28pt] (135,7.9)--(153,7.9);
\fill[hadipoorange!12] (135.5,7.9)
 plot[smooth,samples=70,domain=-1:1] ({144+8.5*\x},{7.9+6.2*exp(-\x*\x/(2*.33*.33))})
 -- (152.5,7.9)--cycle;
\draw[hadipoorange,line width=.7pt,smooth,samples=70,domain=-1:1]
 plot ({144+8.5*\x},{7.9+6.2*exp(-\x*\x/(2*.33*.33))});
\foreach \dx/\lab in {-4.93/A,4.93/B}{
 \node[font=\scriptsize,text=hadipoblue!90!black] at (144+\dx,16.2) {$\lab$};
}

\foreach \dx in {-4.93,4.93}{
 \draw[hadipoblue!55,line width=.35pt,dash pattern=on .8pt off .7pt]
 (144+\dx,7.9)--(144+\dx,13.8);
 \filldraw[fill=white,draw=hadipoblue!80,line width=.4pt]
 (144+\dx,13.8) circle[radius=.65];
}
\fill[hadipoorange!95!black] (144,7.9) circle[radius=.6];
\node[font=\scriptsize,align=center,text=hadipoorange!90!black] at (144,3.4) {possible\\compromise};

% Schematic responses: illustrative possibilities, not empirical densities.
\node[draw=hadipoblue!35,fill=hadipoblue!3,rounded corners=1pt,line width=.35pt,
 minimum width=20.4mm,minimum height=18.1mm,inner sep=0pt] at (166,12) {};
\node[font=\scriptsize\bfseries,text=hadipoblue!90!black] at (166,20.6) {Diffusion};
\draw[black!30,line width=.28pt] (157,7.9)--(175,7.9);
\fill[hadipoblue!12] (157.5,7.9)
 plot[smooth,samples=70,domain=-1:1] ({166+8.5*\x},{7.9+5.7*(exp(-(\x-.58)*(\x-.58)/(2*.18*.18))+exp(-(\x+.58)*(\x+.58)/(2*.18*.18)))})
 -- (174.5,7.9)--cycle;
\draw[hadipoblue,line width=.7pt,smooth,samples=70,domain=-1:1]
 plot ({166+8.5*\x},{7.9+5.7*(exp(-(\x-.58)*(\x-.58)/(2*.18*.18))+exp(-(\x+.58)*(\x+.58)/(2*.18*.18)))});
\foreach \dx/\lab in {-4.93/A,4.93/B}{
 \node[font=\scriptsize,text=hadipoblue!90!black] at (166+\dx,16.2) {$\lab$};
}
\foreach \dx in {-4.93,4.93}{\fill[hadipoblue!95!black] (166+\dx,7.9) circle[radius=.6];}
\node[font=\scriptsize,align=center,text=hadipoblue!90!black] at (166,3.4) {separated\\responses};
\end{tikzpicture}}
  \caption{Overview of \method{}. (a) Heterogeneous robots support an unfixed shared payload. (b) Centralized twin critics guide sequential policy updates through a diffusion-path variational objective, using simulator interaction without demonstrations. (c) Local histories and the team command $\mathbf c$ form $o_i$ for each independent diffusion actor. An embodiment-specific, PPO-pretrained robust policy tracks $a_i$ and remains frozen during high-level coordination learning. (d) The conditional network $f_{\theta_i}$ drives iterative sampling from Gaussian noise, and $T_i$ maps terminal samples to bounded actions. Distribution drawings illustrate representational flexibility. Orange dashed arrows denote training operations.}
  \label{fig:overview}
\end{figure*}
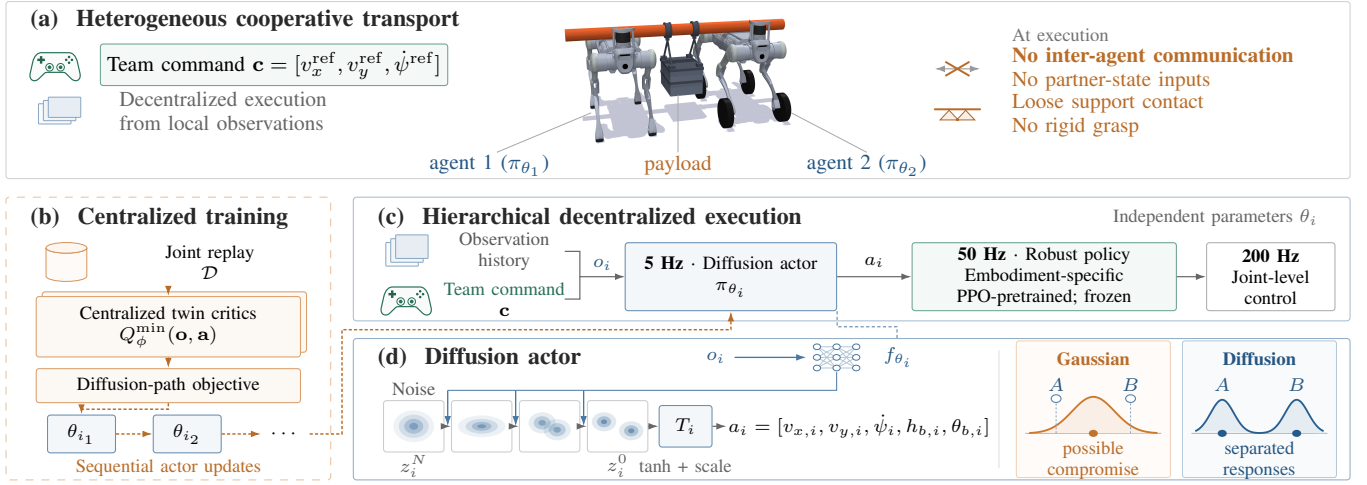

To this end, we propose \method{}, a hierarchical multi-agent reinforcement learning framework for heterogeneous loose-contact transport.
The framework separates team-level coordination from embodiment-specific locomotion (Fig.~\ref{fig:overview}): high-level policies learn how to move and support the shared payload, while robot-specific low-level policies handle robust command tracking.
This division allows coordination to be learned through base-motion commands while retaining controllers tailored to each robot's dynamics and actuation.
Under centralized training with decentralized execution (CTDE), a centralized critic guides independent diffusion actors through team-value estimates.
The high-level policies learn from scratch through reward-driven simulator interaction, without expert demonstrations or imitation pretraining; the pretrained low-level policies remain frozen during coordination learning.
Onboard payload estimates enable each robot to execute from local robot and payload observations, without partner-state inputs or inter-robot messages.

Within this hierarchy, each actor learns a value-guided response to its current teammates.
These responses are shaped by action value and need not be unimodal, motivating diffusion actors that offer greater flexibility than single-Gaussian policies in representing distinct coordination responses~\cite{chi2025diffusion,celik2025dime}.
A variational objective over diffusion paths makes these policies trainable without explicit action-density evaluation.
During training, actors are updated sequentially using the latest teammate policies~\cite{liu2024maximum}.

The contributions of this work are as follows:
\begin{itemize}
    \item \textbf{Team-value-guided diffusion policy optimization for heterogeneous coordination.}
    We extend heterogeneous-agent policy optimization to robot-specific diffusion actors guided by centralized team values.
    Path-based variational learning enables expressive coordination policies to learn from scratch without demonstrations or explicit action-density evaluation.
    \item \textbf{A hierarchical framework for heterogeneous loose-contact transport.}
    We decouple team coordination from embodiment-specific robust control, enabling robots with different dynamics to learn coordinated payload motion and support.
    \item \textbf{Decentralized transport with experimental validation.}
    Local policies coordinate through onboard payload feedback without partner-state inputs or inter-robot messages.
    Simulation comparisons show gains over the evaluated baselines; hardware demonstrates quadrupedal--wheeled-legged and human--robot transport.
\end{itemize}

The remainder of the paper is organized as follows. Section~\ref{sec:related} reviews the related work. Section~\ref{sec:method} presents the main methodological contribution, namely the Diffusion Policy Optimization framework. Section~\ref{sec:system} describes the overall hierarchical control architecture and the onboard perception system used for deployment. Section~\ref{sec:experiments} presents and discusses simulation and experimental results evaluating cooperative transport performance, the ability to represent distinct action responses, and deployment with heterogeneous robot and human partners. Finally, Section~\ref{sec:conclusion} concludes the paper.

\section{Related Work}
\label{sec:related}

Cooperative transport requires coordinating robot motion while maintaining support of a shared payload~\cite{tuci2018review}.
Model-based approaches organize this coupled task through explicit control structures, including task-priority optimization~\cite{koung2021hqp}, layered control~\cite{kim2023layered}, and passive-arm model predictive control~\cite{turrisi2024pacc}.
Learning-based methods complement these designs by acquiring cooperative behavior from interaction or demonstrations.
Reinforcement learning has enabled tether-suspended transport and decentralized carrying~\cite{zhu2023tether,pandit2026decplm}, while demonstration-based learning has produced a joint diffusion policy integrated with whole-body control~\cite{li2026hclm}.

Cooperation with heterogeneous robots or humans additionally requires adaptation to differing motion capabilities and partner behavior.
Physical-interaction feedback provides one basis for this adaptation: force prediction and deformation-agnostic control support human--robot carrying~\cite{vidal2023force,sirintuna2024deformation}, while proprioceptive learning enables human--humanoid coordination~\cite{du2026cola}.
Inferring interaction wrenches from proprioceptive histories further supports the reuse of a transport policy with diverse partners~\cite{cao2026paint}.
A complementary approach is to jointly train interacting policies, as demonstrated in heterogeneous human--robot collaboration with trainable humanoid partners~\cite{zhang2026halypo}.
In this setting, CTDE supports team coordination through centralized value learning while retaining local actor inputs at execution~\cite{lowe2017maddpg}; sequential training updates allow each policy to respond to updated teammates~\cite{liu2024maximum}.

Alongside the coordination mechanism, policy parameterization determines the action distributions that can be represented.
Diffusion policies provide a flexible alternative to single-Gaussian actors, initially demonstrating multimodal manipulation behavior through imitation learning~\cite{chi2025diffusion}.
Demonstration-based generative action modeling has also been extended to vision--language--action systems through diffusion and flow matching~\cite{wen2025diffusionvla,black2025pi0}.
Reward-driven optimization provides a complementary route for improving these policies through interaction, encompassing online diffusion-policy learning~\cite{celik2025dime} and policy-gradient fine-tuning of demonstration-pretrained actors~\cite{ren2025dppo}.
Recent extensions to cooperative multi-agent learning further connect expressive diffusion policies with coordinated exploration~\cite{zhang2026ddpl}.

Taken together, these studies motivate jointly learning expressive, robot-specific coordination policies for heterogeneous loose-contact transport.
We separate team-level coordination from embodiment-specific locomotion and train independent diffusion actors through centralized team-value guidance and variational optimization.
The coordination policies learn from scratch without demonstrations and execute using local observations.

\section{Diffusion Policy Optimization}
\label{sec:method}

Within the sequential heterogeneous-agent policy improvement framework~\cite{liu2024maximum}, we extend diffusion-path entropy regularization~\cite{celik2025dime} to the heterogeneous multi-agent setting and propose Heterogeneous-Agent Diffusion Policy Optimization (\method{}).
The method incorporates per-agent path regularizers into centralized team-value evaluation and jointly optimizes independent diffusion policies conditioned on local observations.

We model cooperative transport as a cooperative Markov game
$\mathcal M=(\mathcal S,\{\mathcal O_i,\mathcal A_i\}_{i=1}^n,P,r,\gamma)$,
where $P(s_{t+1}\mid s_t,\mathbf a_t)$ is the state-transition kernel and $\gamma\in[0,1)$ is the discount factor.
At state $s_t$, robot $i$ uses its local history $o_t^i$ to select a bounded task action $a_t^i\in\mathcal A_i$, and all robots receive the team reward $r(s_t,\mathbf a_t)$.
With $\mathbf o=(o_1,\cdots,o_n)$ denoting the joint local history, the joint policy factorizes as
\begin{equation}
 \pi(\mathbf a\mid\mathbf o)=\prod_{i=1}^n\pi_i(a_i\mid o_i).
 \label{eq:product-policy}
\end{equation}

\subsection{From team rewards to local response distributions}

We seek to maximize the entropy-regularized team return
\begin{equation}
 J_\rho(\pi)=\mathbb E_{\rho,\pi}\!\left[
 \sum_{t\geq0}\gamma^t
 \left(r_t+\sum_i\alpha_i\mathcal H(\pi_i(\cdot\mid o_t^i))\right)
 \right],
 \label{eq:team-maxent}
\end{equation}
balancing task performance and exploration, where $\rho$ is the initial-state distribution, $\mathcal H(\nu)=-\mathbb E_{a\sim\nu}[\log\nu(a)]$ is the differential entropy, and $\alpha_i>0$ controls the entropy regularization of robot $i$.
% We seek to maximize the entropy-regularized team return $J_\rho(\pi)$, balancing task performance and exploration
% \begin{equation}
%  J_\rho(\pi)=\mathbb E_{\rho,\pi}\!\left[
%  \sum_{t\geq0}\gamma^t
%  \left(r_t+\sum_i\alpha_i\mathcal H(\pi_i(\cdot\mid o_t^i))\right)
%  \right],
%  \label{eq:team-maxent}
% \end{equation}
% where $\rho$ is the initial-state distribution and $\alpha_i>0$ controls the entropy regularization of robot $i$.
For a fixed centralized action-value estimate $Q(\mathbf o,\mathbf a)$ and a fixed set of teammate policies $\mu_{-i}$, define the local team-value function for robot $i$ as
\begin{equation}
 U_i^\mu(o_i,a_i)
 =\mathbb E_{\mathcal D,\mu_{-i}}
   [Q(\mathbf o,a_i,\mathbf a_{-i})\mid o_i].
 \label{eq:local-coordinate-energy}
\end{equation}
The expectation marginalizes over the teammates' observations sampled from the replay buffer $\mathcal D$, conditioned on the local observation $o_i$, and over their actions $\mathbf a_{-i}$ sampled from $\mu_{-i}$. Hence, $U_i^\mu(o_i,a_i)$ quantifies the expected team value associated with robot $i$ selecting action $a_i$ from local observation $o_i$, given the current behavior of its teammates.
% For a fixed centralized action-value estimate $Q(\mathbf o,\mathbf a)$ and current teammate policies $\mu_{-i}$, define
% \begin{equation}
%  U_i^\mu(o_i,a_i)
%  =\mathbb E_{\mathcal D,\mu_{-i}}
%    [Q(\mathbf o,a_i,\mathbf a_{-i})\mid o_i].
%  \label{eq:local-coordinate-energy}
% \end{equation}
% The expectation averages teammate histories from replay $\mathcal D$ conditional on $o_i$ and actions sampled from $\mu_{-i}$.
% Thus, $U_i^\mu$ measures the team value of an action available to robot $i$.

Teammates' current entropy terms are constant when optimizing this robot's policy.
For unrestricted action densities, the optimal entropy-regularized response is the Boltzmann distribution~\cite{liu2024maximum}
\begin{equation}
 \begin{split}
 \pi_i^\star(\cdot\mid o_i)
 &=\arg\max_\nu\{\mathbb E_\nu U_i^\mu+\alpha_i\mathcal H(\nu)\},\\
 \pi_i^\star(a_i\mid o_i)
 &=\mathcal B_i^\mu(a_i\mid o_i)
 =\frac{\exp(U_i^\mu(o_i,a_i)/\alpha_i)}{Z_i^\mu(o_i)},
 \end{split}
 \label{eq:coordinate-boltzmann}
\end{equation}
where $Z_i^\mu$ integrates the numerator over $\mathcal A_i$.
This follows by rewriting the maximized objective as
$\alpha_i\log Z_i^\mu-\alpha_iD_{\mathrm{KL}}(\nu\|\mathcal B_i^\mu)$.
Direct sampling from this Boltzmann distribution is generally difficult, and its shape, determined by the action value $U_i^\mu$, may be multimodal.
A single-Gaussian actor is unimodal before action squashing, limiting its ability to approximate such targets.
Motivated by the greater representational flexibility of diffusion policies~\cite{chi2025diffusion,celik2025dime}, we construct agent-specific diffusion samplers and train them to approximate the Boltzmann response using the variational objective below.
Figure~\ref{fig:evidence}(a) illustrates this representational capacity in a controlled toy problem.

\subsection{A neural sampler for bounded actions}

Diffusion models gradually transform complex distributions into simple noise and learn the reverse process~\cite{song2021sde}.
This reverse process serves as a sampler for complex target distributions.
We use this perspective to construct a diffusion policy conditioned on local history.
An Ornstein--Uhlenbeck reference drift provides mean reversion, while a neural correction $f_{\theta_i}$ shapes the generated distribution~\cite{celik2025dime}
\begin{equation}
 \begin{split}
 \mathrm dz_i
 &=\beta_i(t_{\mathrm{sam}})\odot[-\kappa z_i/\sigma_0^2
       +f_{\theta_i}(z_i,t_{\mathrm{sam}},o_i)]\,\mathrm dt_{\mathrm{sam}}\\
 &\quad+\sqrt{2\beta_i(t_{\mathrm{sam}})}\odot\mathrm dW_i(t_{\mathrm{sam}}),
 \end{split}
 \label{eq:diffusion-sde}
\end{equation}
where $\odot$ denotes elementwise multiplication, $t_{\mathrm{sam}}$ is internal sampling time, $W_i$ is standard Brownian motion, and $\kappa$, $\sigma_0$, and the coordinatewise mobility $\beta_i$ are positive.
Choose $0=t_{\mathrm{sam},N}<\cdots<t_{\mathrm{sam},0}$ and set $\lambda_{ik}=\beta_i(t_{\mathrm{sam},k})(t_{\mathrm{sam},k-1}-t_{\mathrm{sam},k})$.
Euler--Maruyama discretization gives, starting from $z_i^N\sim p_N^i=\mathcal N(0,\sigma_0^2I)$,
\begin{equation}
 \begin{split}
 z_i^{k-1}&=z_i^k+\lambda_{ik}\odot
 \left[-\kappa z_i^k/\sigma_0^2+f_{\theta_i}(z_i^k,k,o_i)\right]\\
 &\quad+\sqrt{2\lambda_{ik}}\odot\epsilon_{ik},
 \qquad \epsilon_{ik}\sim\mathcal N(0,I),
 \end{split}
 \label{eq:diffusion-step}
\end{equation}
for $k=N,\cdots,1$, with independent noises; $k$ denotes $t_{\mathrm{sam},k}$ in the network input.
The policy network $f_{\theta_i}$ maps the latent sample, sampling time, and local history to a drift correction.
Its nonlinear dependence on the latent sample permits multimodal endpoint distributions even though every transition is Gaussian.
Only the endpoint is converted to a physical action, $a_i=T_i(z_i^0)$, where $T_i$ applies $\tanh$ elementwise and affinely rescales the result to the action bounds.

\subsection{A computable replacement for action entropy}

Unlike Gaussian policies, diffusion policies generally lack an explicit, tractable action density, making the entropy term in~\eqref{eq:coordinate-boltzmann} difficult to evaluate.
To this end, we extend the variational entropy lower bound in~\cite{celik2025dime} to the heterogeneous multi-agent setting.
The Gaussian transitions in~\eqref{eq:diffusion-step} give the explicit path density
\begin{equation}
 p_i(z_i^{0:N}\mid o_i)
 =p_N^i(z_i^N)\prod_{k=1}^N
 p_{\theta_i}^i(z_i^{k-1}\mid z_i^k,o_i).
 \label{eq:reverse-path}
\end{equation}
To relate this joint density to action entropy, we introduce a tractable reference for the latent path conditioned on its endpoint
$q_i^c(z_i^{1:N}\mid z_i^0)=\prod_{k=1}^Nq_{ik}(z_i^k\mid z_i^{k-1})$, with
\begin{equation}
 q_{ik}(\cdot\mid z)=\mathcal N\!\left(
 z-\lambda_{ik}\odot\kappa z/\sigma_0^2,
 2\operatorname{diag}(\lambda_{ik})\right).
 \label{eq:reference-kernel}
\end{equation}
These normalized kernels retain only the reference drift.
Comparing the reference and generated path densities gives the computable score
\begin{equation}
 \widehat\ell_i=\log q_i^c-\log p_i+\log J_i(z_i^0),
 \label{eq:path-score}
\end{equation}
where $J_i(z)=|\det\nabla T_i(z)|$ and
$\log J_i(z)=\sum_j[\log b_{ij}+\log(1-\tanh^2z_j)]$ accounts for the bounded action transform, with $b_{ij}=(a_{ij}^{\max}-a_{ij}^{\min})/2>0$.
Its conditional expectation
$\mathcal L_i(o_i):=\mathbb E[\widehat\ell_i\mid o_i]\leq\mathcal H(\pi_i(\cdot\mid o_i))$
is the required entropy lower bound; Proposition~1 in Appendix~\ref{sec:analysis} identifies its gap and provides a proof sketch.
Replacing each entropy in~\eqref{eq:team-maxent} by $\mathcal L_i$ defines the surrogate team return $\proxyJ_\rho$ used by both actor and critic.

\subsection{Centralized learning with sequential actor updates}

A centralized neural critic $Q_\phi(\mathbf o,\mathbf a)$ approximates the soft action value associated with the surrogate return $\proxyJ_\rho$.
Let $\bar\phi$ denote target parameters.
For a replay transition $(\mathbf o,\mathbf a,r,\mathbf o',d)$, a one-step Bellman backup with fresh actions and path scores conditioned on next-step histories gives the detached target and critic loss
\begin{equation}
 \begin{split}
 y&=r+\gamma(1-d)\left[
 Q_{\bar\phi}(\mathbf o',\mathbf a')+\sum_i\alpha_i\widehat\ell_i'\right],\\
 \mathcal L_Q&=\mathbb E_{\mathcal D}
 [\operatorname{Huber}(Q_\phi(\mathbf o,\mathbf a)-y)],
 \end{split}
 \label{eq:critic-target}
\end{equation}
respectively, where $d$ indicates termination and the regression uses the recorded joint action.
The Huber loss with threshold $\eta>0$, $\operatorname{Huber}(e)=\tfrac12e^2$ for $|e|\leq\eta$ and $\eta(|e|-\tfrac12\eta)$ otherwise, is quadratic for small temporal-difference errors and linear for large ones, which bounds their gradients and stabilizes critic learning.

Following sequential policy optimization~\cite{liu2024maximum}, we freeze the updated critic and sample an agent order $(i_1,\cdots,i_n)$.
At step $m$, let $\mu^{(m-1)}$ contain the new policies of preceding agents and the old policies of the remaining agents.
The response construction in~\eqref{eq:local-coordinate-energy} uses $Q=Q_\phi$.
Using the entropy lower bound in~\eqref{eq:coordinate-boltzmann} and expressing the resulting maximization as a minimization yields the following actor objective. Approximating its expectations with replay samples and diffusion paths gives
\begin{equation}
 \mathcal L_{\mathrm{actor}}^{i_m}
 =-\mathbb E_{\mathcal D,p_{i_m},\mu_{-i_m}^{(m-1)}}
 \left[Q_\phi(\mathbf o,\mathbf a)+\alpha_{i_m}\widehat\ell_{i_m}\right].
 \label{eq:actor-loss}
\end{equation}
During optimization, gradients are propagated only through the active actor, including its diffusion sampling path, bounded action transformation, and corresponding critic input, while teammate actions are treated as constants. This formulation therefore avoids explicit evaluation of both the actor action density and the normalizing constant of the Boltzmann distribution.

% Replacing entropy in~\eqref{eq:coordinate-boltzmann} by its lower bound, negating the objective, and estimating expectations with replay and sampled paths yields the actor loss
% \begin{equation}
%  \mathcal L_{\mathrm{actor}}^{i_m}
%  =-\mathbb E_{\mathcal D,p_{i_m},\mu_{-i_m}^{(m-1)}}
%  \left[Q_\phi(\mathbf o,\mathbf a)+\alpha_{i_m}\widehat\ell_{i_m}\right].
%  \label{eq:actor-loss}
% \end{equation}
% Gradients pass through the active actor's sampling path, action transform, and critic action input; teammate actions are detached.
% Neither the action density nor the Boltzmann normalizer is evaluated.
Appendix~\ref{sec:analysis} shows that this loss combines Boltzmann response matching with a penalty for the entropy-bound gap.
In implementation, we use two centralized critics and denote their minimum by $Q_\phi^{\min}=\min_{j=1,2}Q_{\phi_j}$.
Algorithm~\ref{alg:training} summarizes one training iteration.

\begin{figure}[!t]
\begin{minipage}{\columnwidth}
\small
\refstepcounter{algorithm}\label{alg:training}
\hrule\smallskip
\textbf{Algorithm \thealgorithm: One HA-DiPO training iteration}\par
\smallskip\hrule
\begin{algorithmic}[1]
 \Require Replay $\mathcal D$, actors $\{\pi_{\theta_i}\}_{i=1}^n$, critics
 $\{Q_{\phi_j},Q_{\bar\phi_j}\}_{j=1}^2$, and \mbox{$\tau\in(0,1]$}.
 \State Collect joint transitions with local policies into $\mathcal D$.
 \State Sample a minibatch $\mathcal B$ from $\mathcal D$.
 \State Sample paths conditioned on $\{o_i'\}$; obtain $\mathbf a'$ and
 $\{\widehat\ell_i'\}$ using~\eqref{eq:diffusion-step} and~\eqref{eq:path-score}.
 \State Compute detached target $y$ using $Q_{\bar\phi}^{\min}$ in~\eqref{eq:critic-target}.
 \State Update $\phi_1,\phi_2$ on $\mathcal B$ using~\eqref{eq:critic-target}.
 \State Hold $\phi_1,\phi_2$ fixed; draw a random order $(i_1,\cdots,i_n)$.
 \State $\mu\gets\prod_{i=1}^n\pi_{\theta_i}$.
 \For{$m=1,\cdots,n$}
  \State $i\gets i_m$.
  \State Sample $z_i^{0:N}\sim p_i(\cdot\mid o_i)$ by reparameterization;
  obtain $a_i,\widehat\ell_i$.
  \State Sample $\mathbf a_{-i}\sim\mu_{-i}(\cdot\mid\mathbf o_{-i})$; stop its gradient.
  \State Update only $\theta_i$ on $\mathcal B$ using $Q_\phi^{\min}$ in~\eqref{eq:actor-loss}.
  \State $\mu_i\gets\pi_{\theta_i}$.
 \EndFor
 \State $\bar\phi_j\gets(1-\tau)\bar\phi_j+\tau\phi_j$, $j=1,2$.
\end{algorithmic}
\hrule
\end{minipage}
\end{figure}
At execution, each robot uses only its local sampler and frozen low-level controller.
Sequential updates, critics, and reference-path evaluation are confined to training.
With fixed temperatures and reference paths, the surrogate return $\proxyJ_\rho$ improves monotonically under exact Markov evaluation and statewise coordinate improvement; Proposition~2 and its proof sketch are given in Appendix~\ref{sec:analysis}.

\section{Hierarchical Control and Onboard Perception}
\label{sec:system}

\subsection{Hierarchical control}
\label{sec:hierarchy}

The hierarchy in Fig.~\ref{fig:overview} separates team-level coordination from robot-specific motion control.
The team command $\mathbf c=[v_x^{\mathrm{ref}},v_y^{\mathrm{ref}},\dot\psi^{\mathrm{ref}}]$ specifies the shared payload's desired planar velocity and yaw rate; robot-specific base-motion commands $a_i$ connect coordination to low-level motion control:
\begin{itemize}
  \item \textbf{High-level coordination policy (5 Hz).}
  Under CTDE, each decentralized diffusion policy maps local robot and payload observations to bounded base-motion commands $a_i$, with dimensions and bounds matched to its robot's capabilities.
  \item \textbf{Robust low-level policy (50 Hz).}
  Using a setup adapted from LocoTouch~\cite{lin2025locotouch}, we pretrain robot-specific policies in simulation with proximal policy optimization (PPO) to track robot $i$'s commanded planar velocity $(v_{x,i},v_{y,i})$, yaw rate $\dot\psi_i$, base height $h_{b,i}$, and pitch $\theta_{b,i}$.
  Dynamics randomization and intermittent pushes promote robustness to model variations and disturbances.
  These policies remain frozen during coordination training.
\end{itemize}
Onboard actor histories contain the team command $\mathbf c$, IMU measurements, previous task action, and local payload estimates, excluding partner poses, velocities, actions, and messages.
During training, the centralized critic receives concatenated joint local histories $\mathbf o$ and task actions $\mathbf a$.
Sections~\ref{sec:perception} and \ref{sec:experiments} describe payload estimation and platform-specific interface dimensions, respectively.

\subsection{Geometry-Constrained Onboard Payload Perception}
\label{sec:perception}

\begin{figure}[t]
  \centering
  \resizebox{\columnwidth}{!}{%
  \begin{tikzpicture}[
   x=1mm,y=.91mm,font=\footnotesize,
   >={Latex[length=1.25mm,width=0.9mm]},
   panel/.style={draw=black!16,rounded corners=1.2pt,fill=white,line width=0.35pt},
   proc/.style={draw=hadipoblue!65,rounded corners=1.1pt,fill=hadipoblue!4,
     align=center,inner sep=0.7pt,font=\scriptsize,line width=0.38pt},
   flow/.style={->,draw=black!48,line width=0.42pt},
  ]
  \path[use as bounding box] (0,12) rectangle (84,56);
  \draw[panel] (0.5,26.5) rectangle (40.2,55.5);
  \draw[panel] (41.3,26.5) rectangle (83.5,55.5);
  \node[fill=hadipoblue!82!black,text=white,rounded corners=0.8pt,
   font=\scriptsize\bfseries,inner xsep=1pt,inner ysep=0.5pt] at (3.5,53.8) {(a)};
  \node[anchor=west,font=\scriptsize\bfseries,text=black!74]
   at (6,53.8) {Example layout};
  \node[fill=hadipogreen!85!black,text=white,rounded corners=0.8pt,
   font=\scriptsize\bfseries,inner xsep=1pt,inner ysep=0.5pt] at (44,53.8) {(b)};
  \node[anchor=west,font=\scriptsize\bfseries,text=black!74]
   at (46.5,53.8) {Geometric model};

  \node[inner sep=0pt] at (20.35,39.5)
   {\includegraphics[width=38.4mm]{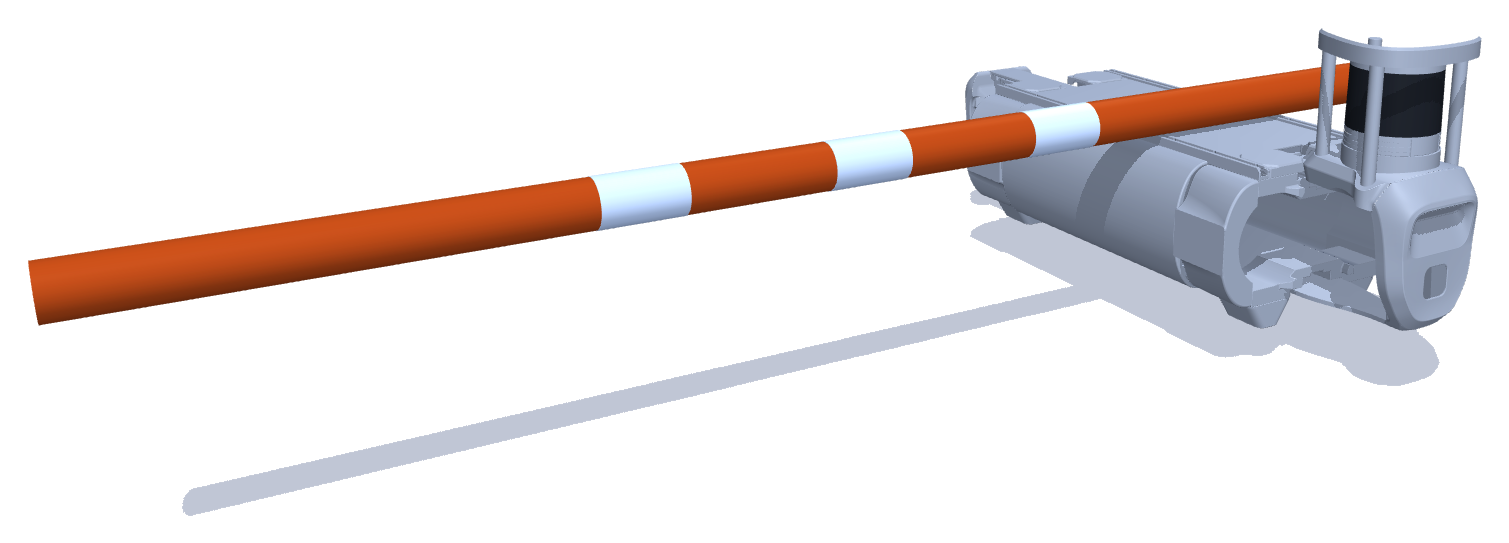}};
  \coordinate (bandminus) at (17.6,41.3);
  \coordinate (bandcenter) at (23.4,42.4);
  \coordinate (bandplus) at (28.4,43.2);
  \coordinate (lidar) at (36.2,43.8);
  \foreach \target in {bandminus,bandcenter,bandplus}{
   \draw[hadipoblue!60,densely dotted,line width=0.4pt] (lidar) -- (\target);
  }
  \node[align=center,font=\scriptsize,text=black!72,inner sep=1pt]
   (bands) at (18.7,49.2) {retroreflective bands};
  \draw[->,black!45,line width=0.33pt] (bands.south west) -- (bandminus);
  \draw[->,black!45,line width=0.33pt] (bands.south) -- (bandcenter);
  \draw[->,black!45,line width=0.33pt] (bands.south east) -- (bandplus);
  \node[draw=hadipoblue!55,fill=white,rounded corners=0.7pt,
   inner sep=1pt,font=\scriptsize,text=hadipoblue!90!black]
   (lidarlabel) at (9.6,33) {3-D LiDAR};
  \draw[->,hadipoblue!65,line width=0.36pt]
   (lidarlabel.east) to[out=5,in=270] (lidar);
  \node[font=\scriptsize,text=black!58] at (20.3,28.9) {three bands: $-\delta,\ 0,\ +\delta$};

  \begin{scope}[yshift=-2.4mm]
  % Equally spaced observed cluster centres; no inferred marker in this panel.
  \coordinate (pm) at (48.6,40.5);
  \coordinate (pc) at (62.1,43.4);
  \coordinate (pp) at (75.6,46.3);
  \draw[hadipoorange!55,line width=1.15pt,line cap=round] (45.7,39.88) -- (78.5,46.92);
  \draw[->,hadipoorange!90!black,line width=0.65pt]
   (pc) -- (80,47.25) node[above,font=\footnotesize,inner sep=1pt] {$\hat u$};
  \foreach \cx/\cy in {48.6/40.5,62.1/43.4,75.6/46.3}{
   \begin{scope}[shift={(\cx,\cy)}]
    \draw[hadipoblue!20,fill=hadipoblue!4] (0,0) ellipse (3.5 and 2.45);
    \foreach \dx/\dy in {-2.5/-.7,-2.1/.6,-1.6/1.5,-1.4/-1.6,
      -1/-.2,-.6/.8,-.2/1.8,.1/-1.2,.4/.3,.8/1.3,1.1/-.8,
      1.5/.5,1.9/1.4,2.3/-.4,2.7/.7}{
     \fill[hadipoblue!76] (\dx,\dy) circle[radius=0.25];
    }
    \draw[hadipogreen!85!black,fill=white,line width=0.65pt] (0,0) circle[radius=0.75];
    \fill[hadipogreen!85!black] (0,0) circle[radius=0.23];
   \end{scope}
  }
  \node[font=\footnotesize,text=black!74] at (48.6,45.3) {$\hat p_-$};
  \node[font=\footnotesize,text=black!74] at (62.1,48.5) {$\hat p_0$};
  \node[font=\footnotesize,text=black!74] at (75.6,51.5) {$\hat p_+$};
  \foreach \x/\y in {45/49,54.4/51.2,56.5/38.8,67.4/52.6,70.3/39.1,80.3/41}{
   \fill[black!19] (\x,\y) circle[radius=0.22];
  }
  \foreach \x/\y in {48.6/38,62.1/40.9,75.6/43.8}{
   \draw[black!25,densely dotted,line width=0.3pt] (\x,34.6) -- (\x,\y);
  }
  \draw[<->,black!48,line width=0.35pt] (48.6,35) --
   node[fill=white,inner sep=0.6pt,font=\scriptsize] {$\delta$} (62.1,35);
  \draw[<->,black!48,line width=0.35pt] (62.1,35) --
   node[fill=white,inner sep=0.6pt,font=\scriptsize] {$\delta$} (75.6,35);
  \node[font=\scriptsize,text=black!62] at (62.4,31.9) {$\delta=0.25$ m\quad $L=2\delta$};

  \end{scope}

  % Onboard processing pipeline.
  \node[anchor=west,font=\scriptsize\bfseries,text=black!62]
   at (1,23.6) {Onboard estimation};
  \draw[black!14,line width=0.32pt] (30,23.6) -- (83,23.6);
  \node[proc,minimum width=18.4mm,minimum height=7mm] (gate) at (10.3,16.8)
   {range / ROI\\intensity gate};
  \node[proc,minimum width=19.4mm,minimum height=7mm] (cluster) at (31.2,16.8)
   {3-D DBSCAN};
  \node[proc,draw=hadipogreen!80!black,fill=hadipogreen!5,
   minimum width=21.5mm,minimum height=7mm] (assoc) at (53.6,16.8)
   {geometry + time\\association};
  \node[proc,draw=hadipogreen!80!black,fill=hadipogreen!5,
   minimum width=16.8mm,minimum height=7mm] (kf) at (75.1,16.8)
   {constrained\\CV Kalman};
  \draw[flow] (gate) -- (cluster);
  \draw[flow] (cluster) -- (assoc);
  \draw[flow] (assoc) -- (kf);
  \end{tikzpicture}}
  \par\vspace{0.3mm}
  {\scriptsize\bfseries Real-robot LiDAR measurements\par}
  \vspace{0.2mm}
  \includegraphics[width=\columnwidth]{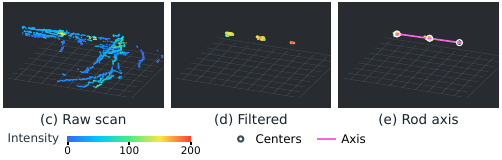}
  \caption{Geometry-constrained onboard payload perception. (a), (b) Example three-band layout and geometry, with the estimation pipeline below. (c)--(e) Real-robot LiDAR measurements from one static scan: raw points, filtered returns, and cluster centers (white circles) with the constrained rod-axis estimate (magenta).}
  \label{fig:rod-perception}
\end{figure}
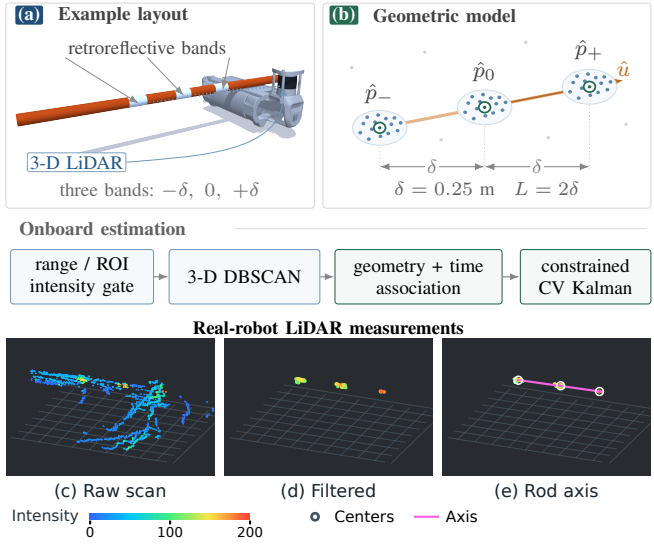

For deployment, we use only onboard 3-D LiDAR and IMU measurements to estimate the payload center and axis for local coordination.
Figure~\ref{fig:rod-perception} illustrates a three-band example; hardware deployment uses five reflective markers with the corresponding observation model and geometric constraints.

% Combined simulation and hardware gallery, six panels per row.
\begin{figure*}[t]
  \centering
  {\raggedright\footnotesize\bfseries Sim-to-sim: Isaac Lab $\rightarrow$ MuJoCo\par}\vspace{0.3mm}
  \begin{minipage}[t]{0.16\textwidth}
    \centering
    \begin{tikzpicture}
      \node[anchor=south west,inner sep=0] (panel) at (0,0) {\includegraphics[width=\linewidth,trim=0 0 0 80bp,clip]{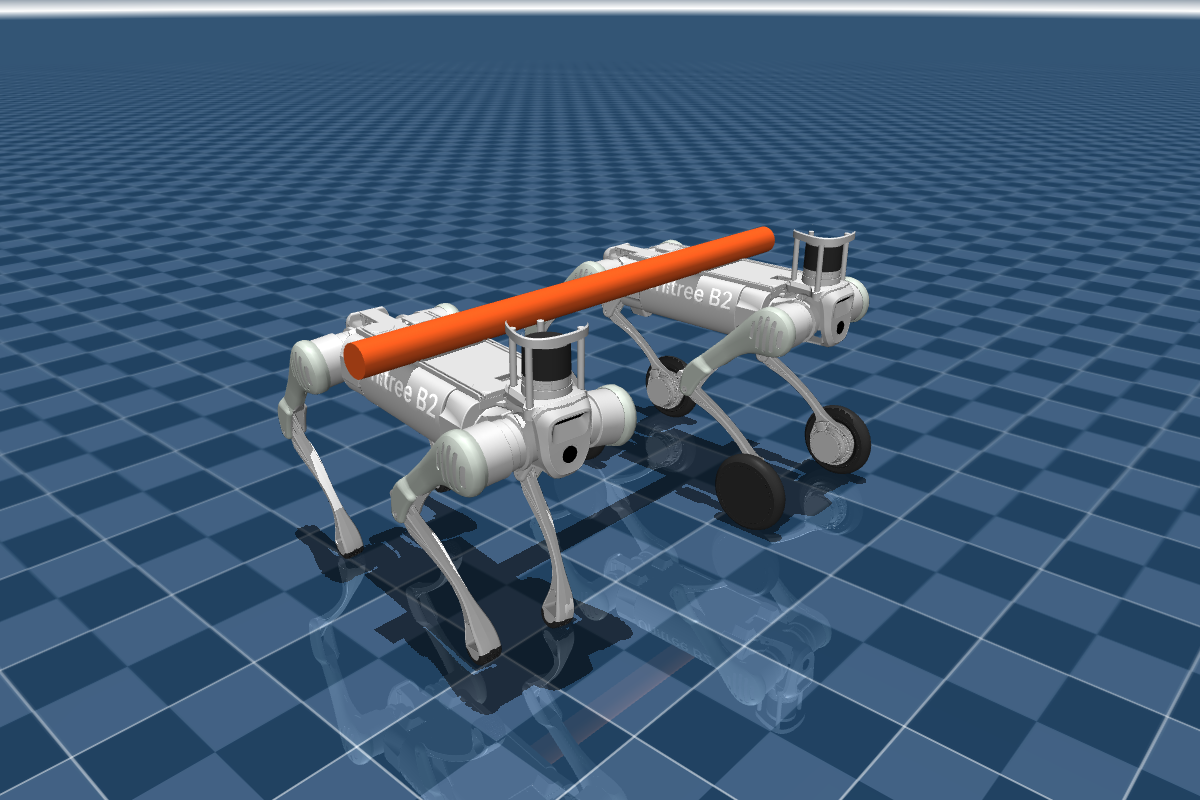}};
      \begin{scope}[x={(panel.south east)},y={(panel.north west)}]
        \draw[cyan!35!white,line width=0.8pt,-{Stealth[length=1.3mm,width=1mm]}] (0.10,0.90)--(0.31,0.79);
      \end{scope}
    \end{tikzpicture}
    \par\vspace{0.3mm}\footnotesize (a) Forward
  \end{minipage}\hfill
  \begin{minipage}[t]{0.16\textwidth}
    \centering
    \begin{tikzpicture}
      \node[anchor=south west,inner sep=0] (panel) at (0,0) {\includegraphics[width=\linewidth,trim=0 0 0 80bp,clip]{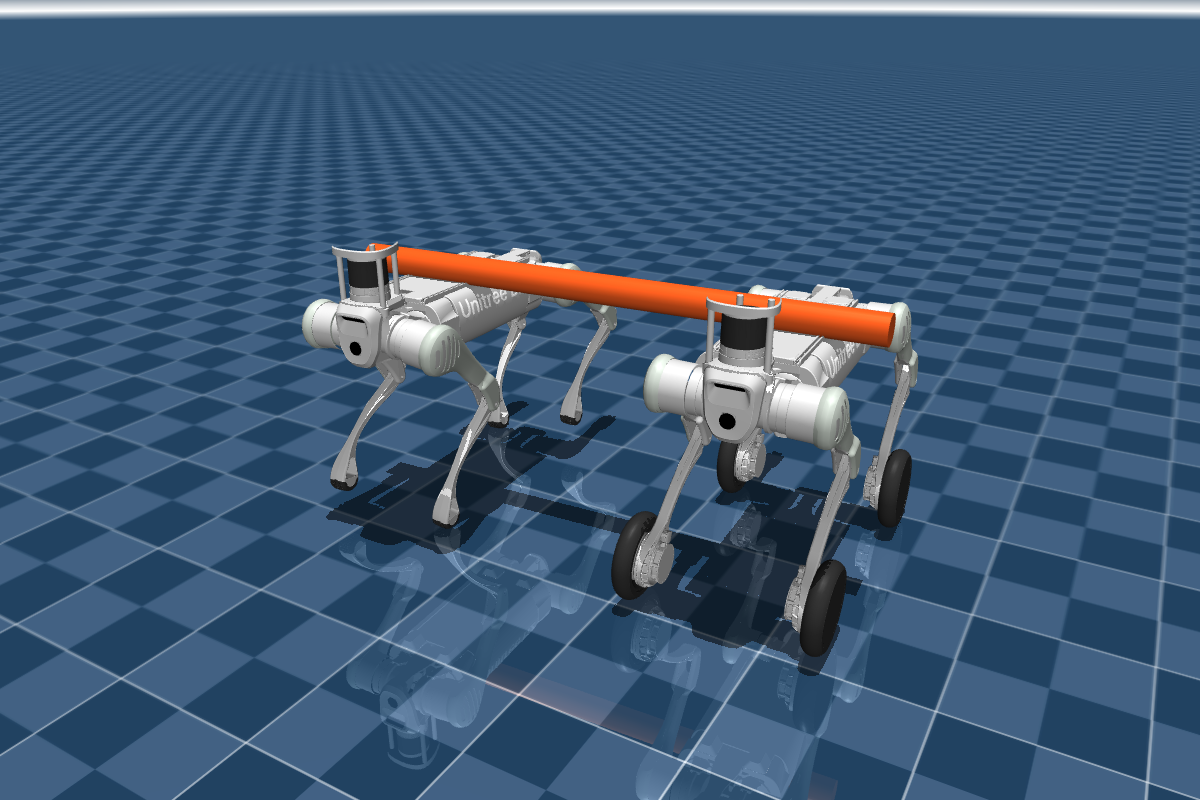}};
      \begin{scope}[x={(panel.south east)},y={(panel.north west)}]
        \draw[cyan!35!white,line width=0.8pt,-{Stealth[length=1.3mm,width=1mm]}] (0.10,0.77)--(0.29,0.96);
      \end{scope}
    \end{tikzpicture}
    \par\vspace{0.3mm}\footnotesize (b) Backward
  \end{minipage}\hfill
  \begin{minipage}[t]{0.16\textwidth}
    \centering
    \begin{tikzpicture}
      \node[anchor=south west,inner sep=0] (panel) at (0,0) {\includegraphics[width=\linewidth,trim=0 0 0 80bp,clip]{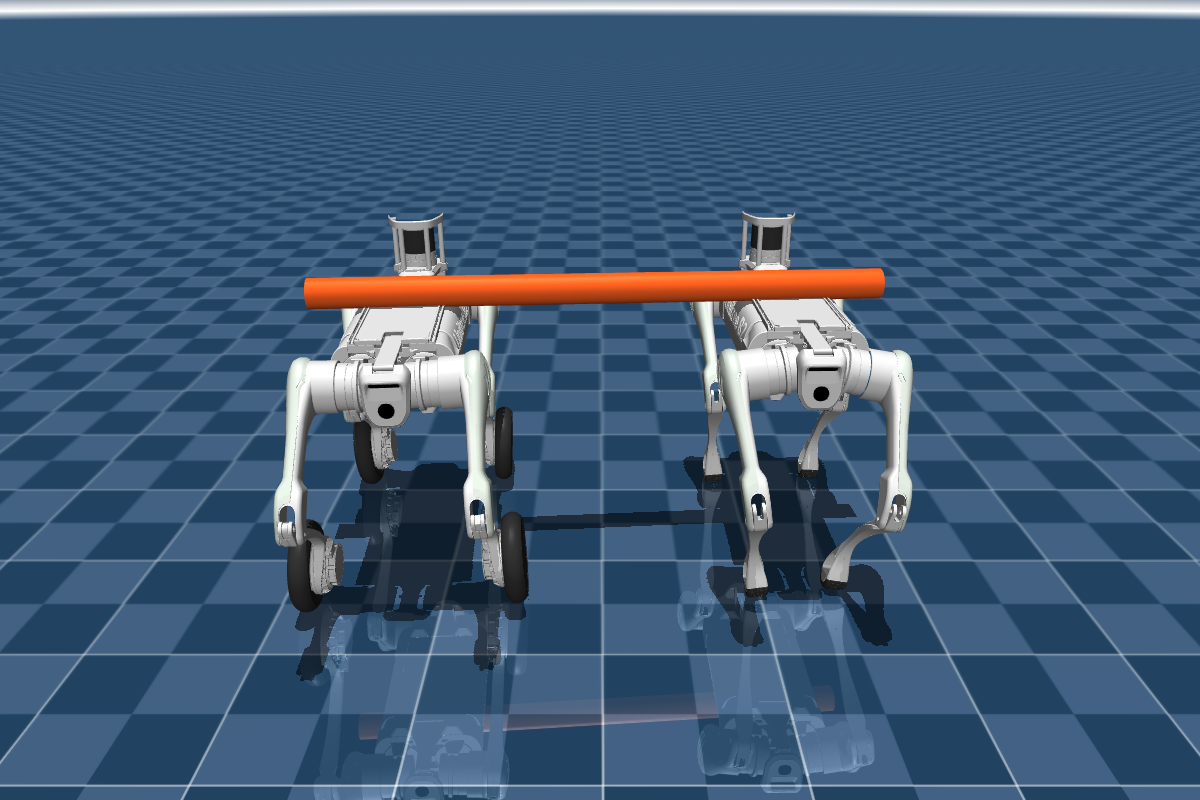}};
      \begin{scope}[x={(panel.south east)},y={(panel.north west)}]
        \draw[cyan!35!white,line width=0.8pt,-{Stealth[length=1.3mm,width=1mm]}] (0.32,0.85)--(0.09,0.85);
      \end{scope}
    \end{tikzpicture}
    \par\vspace{0.3mm}\footnotesize (c) Lateral
  \end{minipage}\hfill
  \begin{minipage}[t]{0.16\textwidth}
    \centering
    \begin{tikzpicture}
      \node[anchor=south west,inner sep=0] (panel) at (0,0) {\includegraphics[width=\linewidth,trim=0 0 0 80bp,clip]{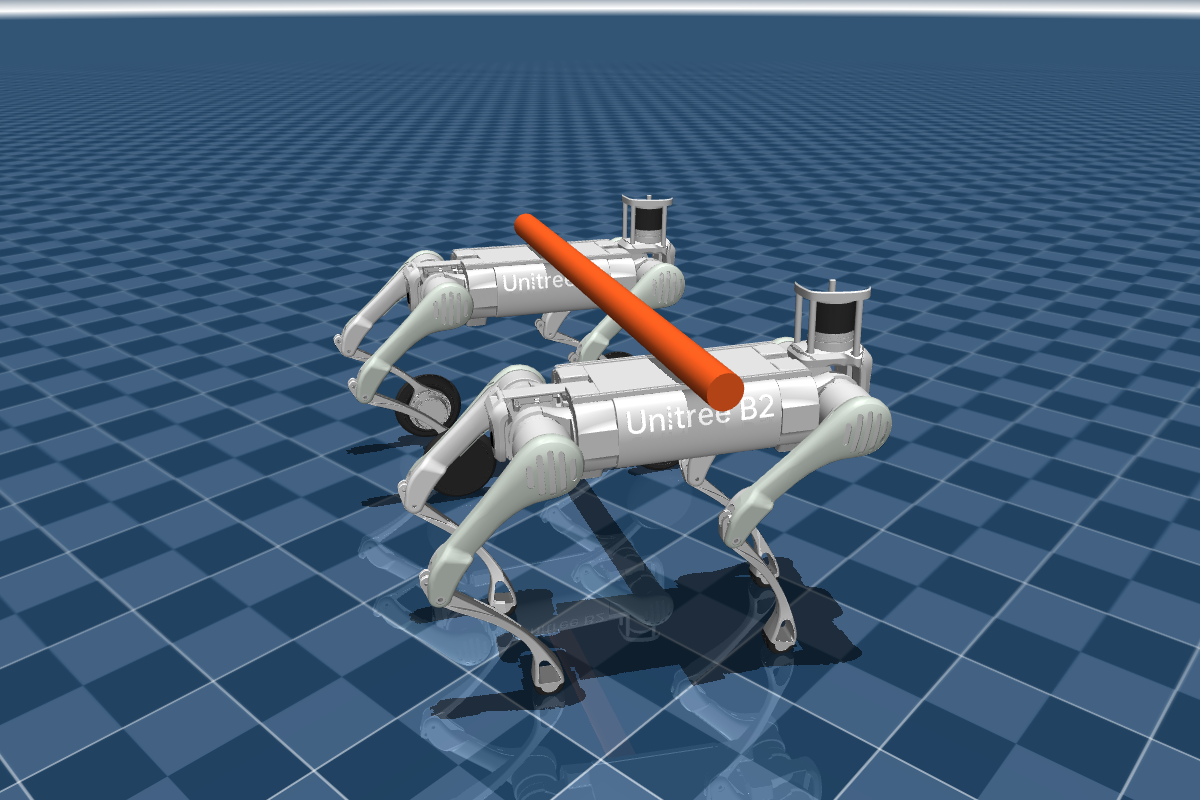}};
      \begin{scope}[x={(panel.south east)},y={(panel.north west)}]
        \draw[cyan!35!white,line width=0.8pt,-{Stealth[length=1.3mm,width=1mm]}] (0.27,0.85) arc[start angle=0,end angle=290,x radius=0.065,y radius=0.0975];
      \end{scope}
    \end{tikzpicture}
    \par\vspace{0.3mm}\footnotesize (d) Left turn
  \end{minipage}\hfill
  \begin{minipage}[t]{0.16\textwidth}
    \centering
    \begin{tikzpicture}
      \node[anchor=south west,inner sep=0] (panel) at (0,0) {\includegraphics[width=\linewidth,trim=0 0 0 80bp,clip]{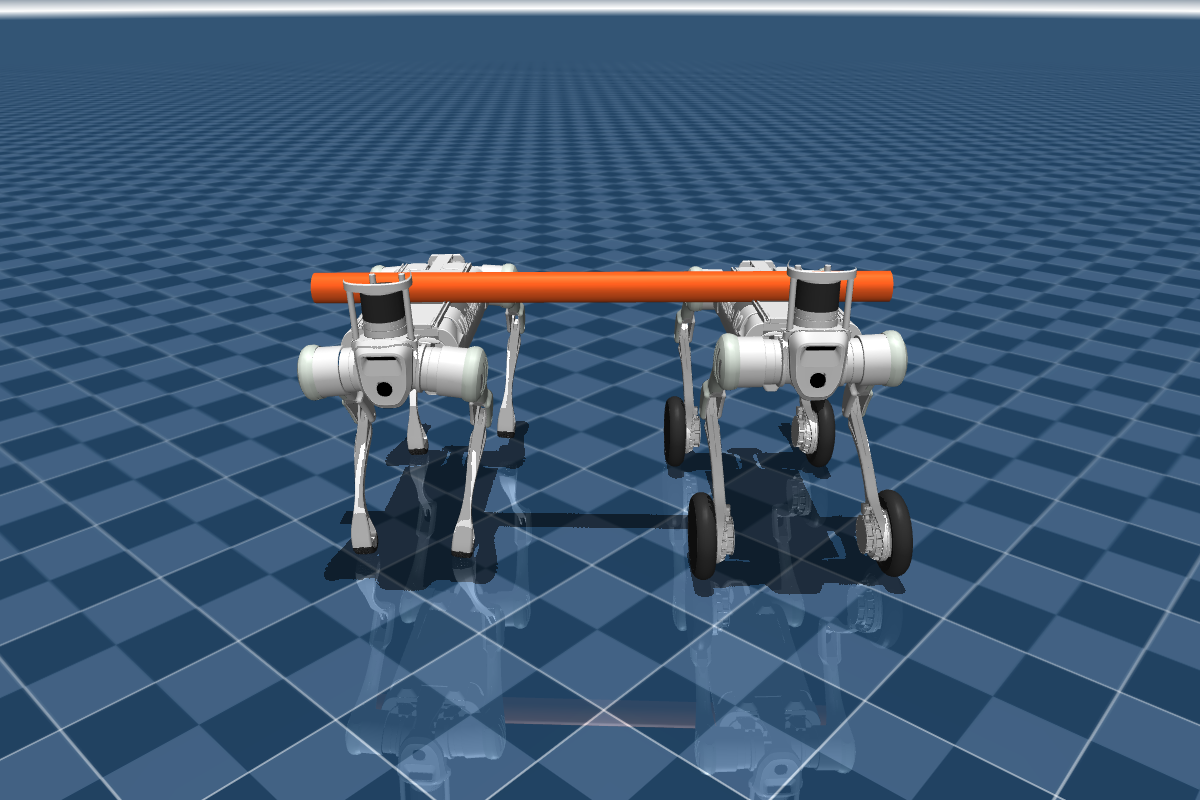}};
      \begin{scope}[x={(panel.south east)},y={(panel.north west)}]
        \draw[cyan!35!white,line width=0.8pt,-{Stealth[length=1.3mm,width=1mm]}] (0.27,0.85) arc[start angle=0,end angle=-290,x radius=0.065,y radius=0.0975];
      \end{scope}
    \end{tikzpicture}
    \par\vspace{0.3mm}\footnotesize (e) Right turn
  \end{minipage}\hfill
  \begin{minipage}[t]{0.16\textwidth}
    \centering
    \begin{tikzpicture}
      \node[anchor=south west,inner sep=0] (panel) at (0,0) {\includegraphics[width=\linewidth,trim=0 0 0 80bp,clip]{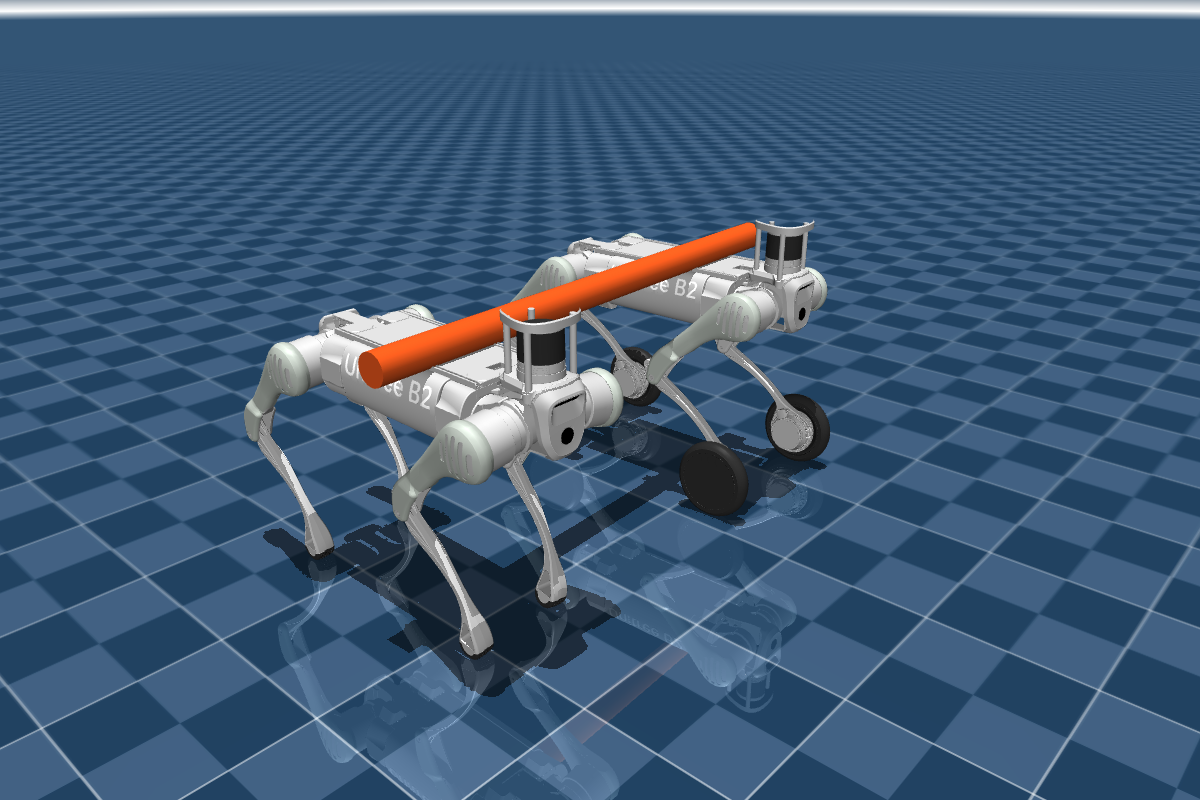}};
      \begin{scope}[x={(panel.south east)},y={(panel.north west)}]
        \draw[orange,line width=1pt,-{Stealth[length=1.5mm,width=1.15mm]}] (0.29,0.76)--(0.47,0.665);
      \end{scope}
    \end{tikzpicture}
    \par\vspace{0.3mm}\footnotesize (f) Disturbance
  \end{minipage}
  \par\vspace{0.9mm}
  {\raggedright\footnotesize\bfseries Real-world experiments\par}\vspace{0.3mm}
  \begin{minipage}[t]{0.16\textwidth}
    \centering
    \includegraphics[width=\linewidth]{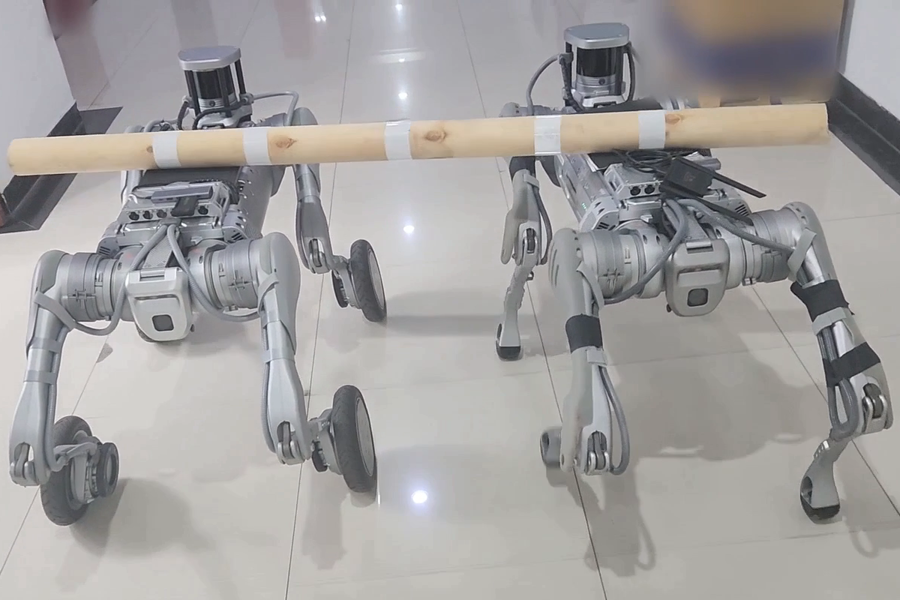}
    \par\vspace{0.3mm}\footnotesize (g) Straight
  \end{minipage}\hfill
  \begin{minipage}[t]{0.16\textwidth}
    \centering
    \includegraphics[width=\linewidth]{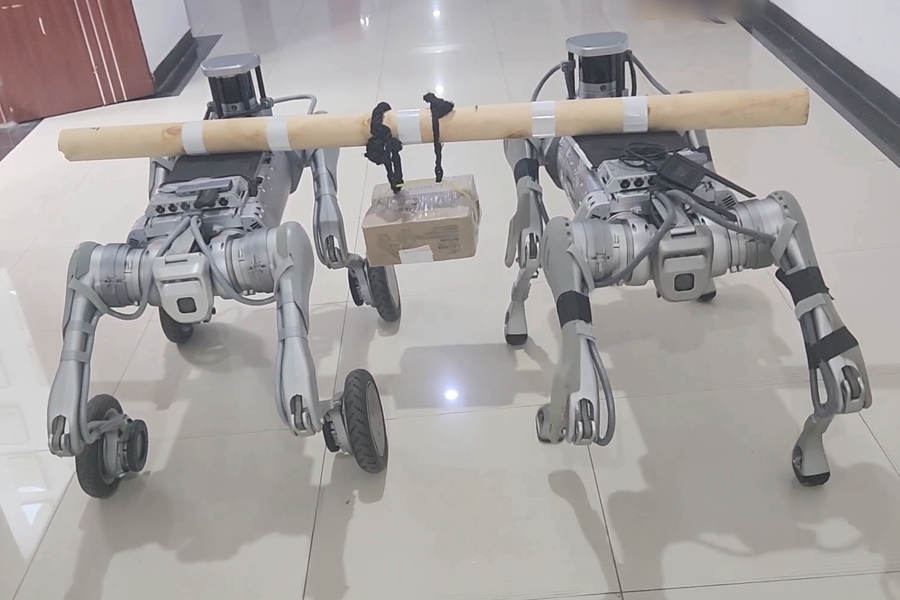}
    \par\vspace{0.3mm}\footnotesize (h) Loaded straight
  \end{minipage}\hfill
  \begin{minipage}[t]{0.16\textwidth}
    \centering
    \includegraphics[width=\linewidth]{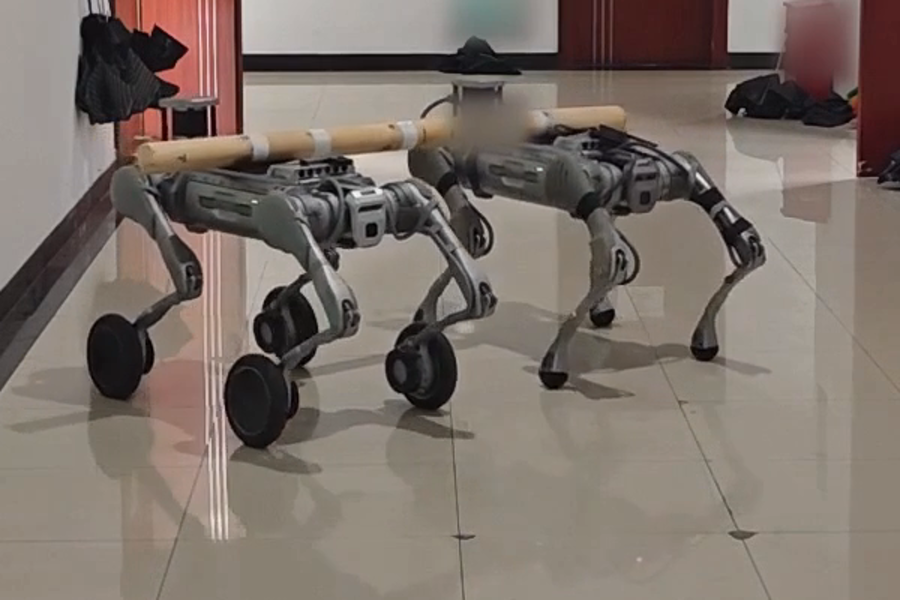}
    \par\vspace{0.3mm}\footnotesize (i) Turn
  \end{minipage}\hfill
  \begin{minipage}[t]{0.16\textwidth}
    \centering
    \includegraphics[width=\linewidth]{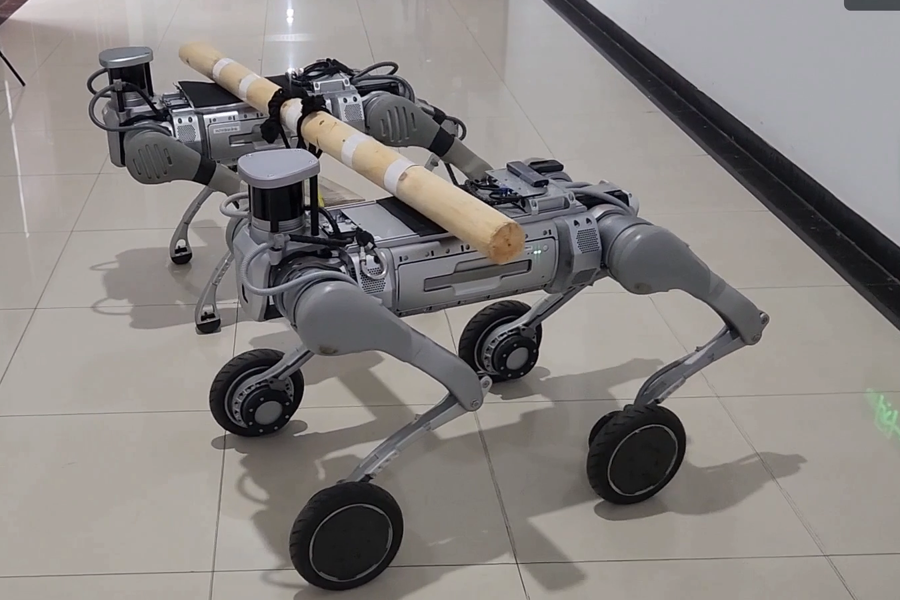}
    \par\vspace{0.3mm}\footnotesize (j) Loaded turn
  \end{minipage}\hfill
  \begin{minipage}[t]{0.16\textwidth}
    \centering
    \includegraphics[width=\linewidth]{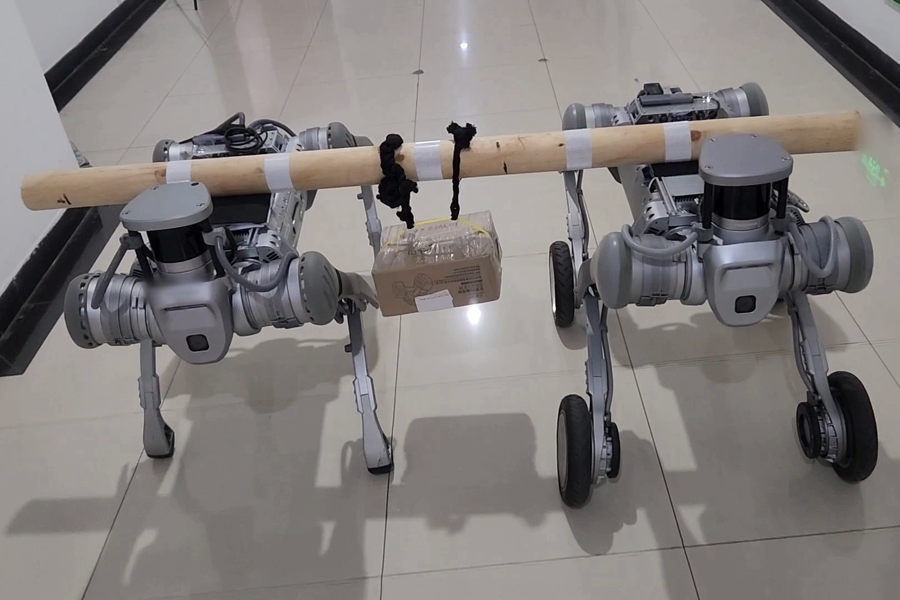}
    \par\vspace{0.3mm}\footnotesize (k) After turn
  \end{minipage}\hfill
  \begin{minipage}[t]{0.16\textwidth}
    \centering
    \includegraphics[width=\linewidth]{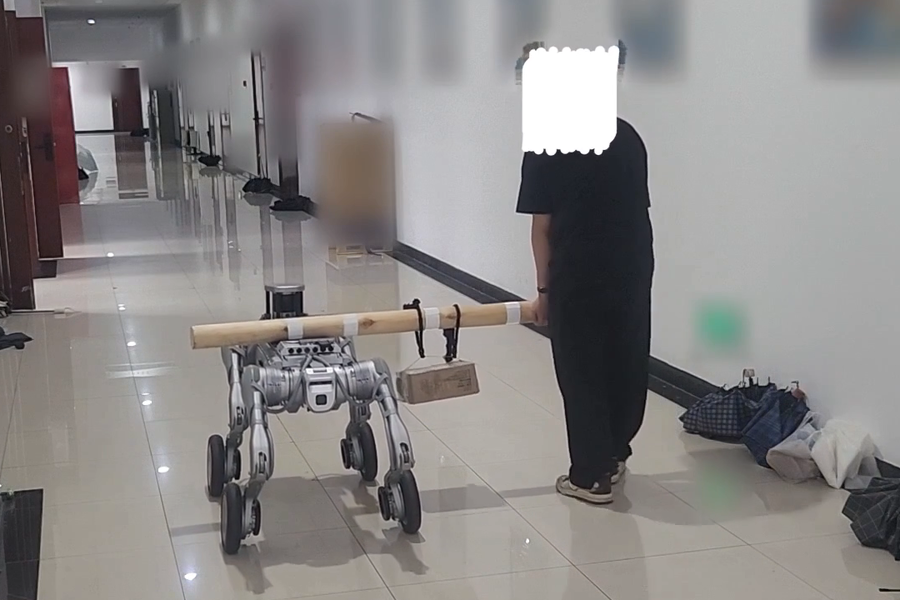}
    \par\vspace{0.3mm}\footnotesize (l) Human--robot
  \end{minipage}
  \caption{Cooperative transport in simulation and hardware. (a)--(f) Direct Isaac Lab-to-MuJoCo policy transfer for B2--B2W translation, turning, and disturbance tests; cyan/orange arrows mark motion commands/disturbances. (g), (h) Real-world straight travel without/with an additional payload. (i) Turning without an additional payload. (j), (k) Two configurations during one loaded turn. (l) Human--robot transport with a B2W.}
  \label{fig:transport-gallery}
\end{figure*}
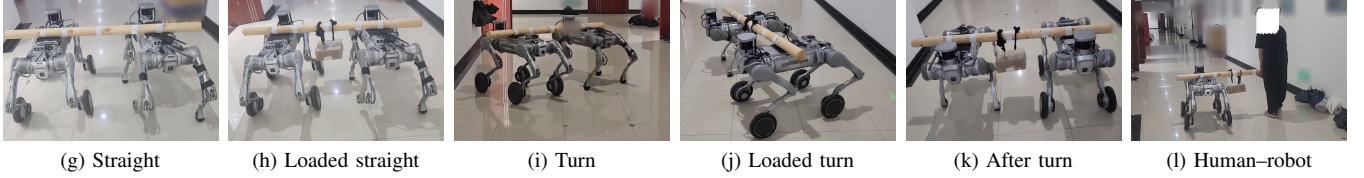

\subsubsection{Marker geometry and association}
For three retroreflective bands spaced by $\delta$, the marker positions encode the rod center $p_0$ and unit axis $u$, that is, $p_\pm=p_0\pm\delta u$.
% \begin{equation}
%  p_-=c-du,\qquad p_c=c,\qquad p_+=c+du.
%  \label{eq:marker-geometry}
% \end{equation}
The outer-marker separation is $L=2\delta$.
Each LiDAR scan is first gated by range, a region of interest (ROI), and return intensity, retaining mainly returns from the retroreflective bands.
The remaining points are grouped using 3-D DBSCAN~\cite{ester1996dbscan}.
The resulting cluster centers serve as marker candidates and are associated with the rod markers using the known geometry and temporal predictions.

\subsubsection{Constrained Kalman filtering}
A constrained Kalman filter tracks outer-marker positions and velocities relative to the LiDAR, with state
$x=[p_-^\top,p_+^\top,v_-^\top,v_+^\top]^\top\in\mathbb R^{12}$.
Using a constant-velocity (CV) approximation over sampling interval $\Delta t_k$, it predicts
$x_k^-=F_kx_{k-1}$, $P_k^-=F_kP_{k-1}F_k^\top+Q_k$,
% \begin{equation}
%  x_k^-=F_kx_{k-1},\qquad P_k^-=F_kP_{k-1}F_k^\top+Q_k,
%  \label{eq:rod-prediction}
% \end{equation}
where $F_k$ advances each position by $\Delta t_kv_\pm$ while leaving velocities unchanged, and $P_k,Q_k$ are the state and process-noise covariances.
The midpoint geometry gives a linear observation model for measured centers stacked in the order $(-,0,+)$,
\begin{equation}
 z_k=H_3x_k+\nu_k,\qquad
 H_3=\begin{bmatrix}
 I&0&0&0\\ \tfrac12 I&\tfrac12 I&0&0\\0&I&0&0
 \end{bmatrix},
 \label{eq:rod-observation}
\end{equation}
where $I$ is the $3\times3$ identity matrix and $\nu_k$ has covariance $R_k$, set by cluster dispersion.
Selecting detected, associated markers gives $H_k$ from $H_3$ and the corresponding $z_k,R_k$, yielding the correction
\begin{equation}
 \begin{split}
 K_k&=P_k^-H_k^\top(H_kP_k^-H_k^\top+R_k)^{-1},\\
 x_k&=x_k^-+K_k(z_k-H_kx_k^-).
 \end{split}
 \label{eq:rod-update}
\end{equation}
Enforcing $\lVert p_+-p_-\rVert=L$ after prediction and correction preserves the fixed geometry.
The outputs are the center $\hat p_0=(\hat p_-+\hat p_+)/2$ and axis $\hat u=(\hat p_+-\hat p_-)/\lVert\hat p_+-\hat p_-\rVert$.
Sensor extrinsics and IMU attitude transform these estimates into the robot's gravity-aligned frame to provide local payload feedback to its coordination policy.

\section{Experimental Evaluation}
\label{sec:experiments}
\label{sec:results}

%\textit{Implementation details:}

The experimental evaluation combines quantitative simulation benchmarks, controlled policy-representation analysis, and real-world deployment to address three main questions. First, does \method{} improve cooperative transport performance compared with existing multi-agent reinforcement learning baselines? Second, do diffusion actors provide a representational advantage over Gaussian policies when coordination requires distinct high-value action responses? Third, can the learned coordination behavior be deployed with heterogeneous robot and human partners while preserving decentralized execution from local observations?

For all reported evaluations, \method{} uses three 256-unit hidden layers, GELU activations, and Fourier time embeddings in each diffusion network, with $N=8$ sampling steps. Each centralized critic is a ReLU MLP with hidden widths $(256,256,128)$, trained with Huber threshold $\eta=10$. Actors and critics are optimized with Adam, while the entropy temperatures are adjusted automatically during training. Training is performed on NVIDIA A40 GPUs using a server equipped with two Intel Xeon Gold 6346 CPUs and 1~TB of RAM.

\subsection{Cooperative Transport Performance}
\label{sec:transport-performance}

% Compact single-column learning curves.
\begin{figure}[t!]
  \centering
  \includegraphics[width=\columnwidth,trim=0 0 0 4bp,clip]{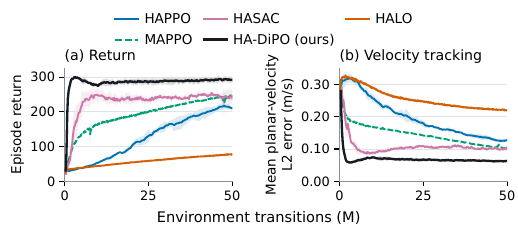}
  \caption{Learning curves for heterogeneous B2--B2W transport: (a) episode return and (b) mean planar-velocity L2 error. Curves show the mean and shading indicates $\pm$ one standard deviation across three seeds, each trained for approximately 50 million environment transitions.}
  \label{fig:learning-curves}
  \label{fig:result-reserve}
\end{figure}

Isaac Lab-trained policies transfer directly to MuJoCo for forward, backward, lateral, turning, and payload-disturbance tests (Fig.~\ref{fig:transport-gallery}(a)--(f)).
Training and independent evaluation use Isaac Lab~\cite{mittal2025isaaclabshort}, with a Unitree B2 quadruped and B2W wheeled-legged robot supporting an unfixed cylindrical payload.
Other elongated cross-sections can be addressed by adapting perception and training settings.
Each coordination actor receives six 28-D frames (168 dimensions) of simulator-provided robot and payload observations, excludes teammate states, and outputs the 5-D command $a_i=[v_{x,i},v_{y,i},\dot\psi_i,h_{b,i},\theta_{b,i}]$ through the hierarchy (Section~\ref{sec:hierarchy}).
The two \method{} actors share architecture, observation--action interfaces, and bounds but have independent parameters; platform heterogeneity arises from distinct actuation and locomotion dynamics.
The frozen low-level policies output 12 leg-joint position residuals for B2 and 16 commands for B2W: 12 leg-joint position residuals and four wheel velocities.
Physics and joint-level control run at 200 Hz.
Training randomizes payload mass, friction, robot separation, and overhang.
Team commands span $v_x^{\mathrm{ref}}\in[-0.6,0.6]$ and $v_y^{\mathrm{ref}}\in[-0.3,0.3]$ m/s, with reference planar speed bounded by 0.6 m/s throughout training and no command curriculum.

All robots share a team reward
\begin{equation}
 r_t=w_{\mathrm{trk}}r_{\mathrm{trk}}
 -w_{\mathrm{sup}}g_{\mathrm{sup}}
 -w_{\mathrm{co}}g_{\mathrm{co}}
 -w_{\mathrm{reg}}g_{\mathrm{reg}},
 \label{eq:transport-reward}
\end{equation}
encouraging payload-motion tracking and stable support, where $r_{\mathrm{trk}}$ rewards tracking of the payload-motion reference $\mathbf c$,
$g_{\mathrm{sup}}$ penalizes payload-center offsets and support-height imbalance,
$g_{\mathrm{co}}$ penalizes relative-position errors between robots and inconsistent support-point motion,
and $g_{\mathrm{reg}}$ regularizes body-pitch commands.
The $w$ coefficients are the corresponding weights.\par

The comparison includes MAPPO~\cite{yu2022mappo}, HAPPO~\cite{kuba2022trust}, HASAC~\cite{liu2024maximum}, and HALO~\cite{zhang2026halypo}.
Within the established sequential optimization framework~\cite{liu2024maximum}, HASAC serves as the Gaussian-policy ablation reference for assessing diffusion-based policy learning.
All methods use identical task observations, rewards, action bounds, and frozen low-level policies, with a common budget of approximately 50 million environment transitions per seed and at least three independent seeds.
Evaluation uses matched initial conditions, a common policy-selection rule, and identical performance criteria.
We use a shared actor for MAPPO, consistent with its original configuration~\cite{yu2022mappo}, as parameter sharing outperformed separate actors in our comparisons.
HALO is adapted to the common transport task.

% Queue the wide policy analysis near the experimental discussion.
\begin{figure*}[t]
  \centering
  \begin{minipage}[t]{0.42\textwidth}
    \centering
    \includegraphics[width=\linewidth,trim=0 4bp 0 0,clip]{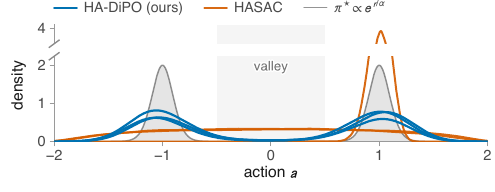}
    \small (a) Bimodal toy problem
  \end{minipage}\hfill
  \begin{minipage}[t]{0.56\textwidth}
    \centering
    \includegraphics[width=\linewidth,trim=0 3bp 0 5bp,clip]{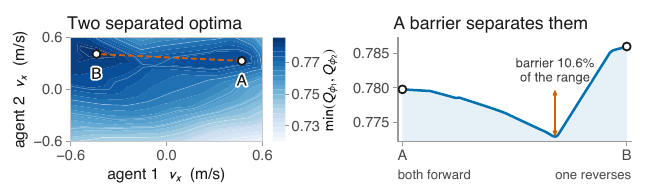}
    \small (b) Transport value slice
  \end{minipage}
  \caption{Policy representation and transport value structure. (a) Three-seed action densities on the bimodal bandit; gray denotes the Boltzmann target. (b) Forward-velocity slice of the learned clipped double-Q function at one visited transport state, with other actions fixed. Translation and pivoting form distinct maxima separated by lower values along the indicated straight segment.}
  \label{fig:evidence}
\end{figure*}

Figure~\ref{fig:learning-curves} compares return and planar-velocity tracking across methods.
\method{} combines rapid improvement early in training with the highest final mean return and the lowest mean planar-velocity error among the compared methods.
At the final recorded point, its planar-velocity error is $0.0634\pm0.0019$ m/s (mean $\pm$ standard deviation across seeds), corresponding to a 37.3\% reduction in mean error relative to MAPPO, the strongest baseline on this metric.

Table~\ref{tab:main-results} tests seven nonzero commands: forward/backward, left/right translation, left/right pure turning, and combined motion.
Deterministic evaluation uses 64 matched initializations per command (448 episodes per seed).
For each approximately 50M-step training run, we evaluate the policy with the highest recorded mean training reward.
Episodes end at 19.8 s or upon robot instability, payload-support loss, protected-body intrusion, or excessive tracking error.

Survival denotes reaching the time limit without failure.
Velocity-tracking success additionally requires $\geq90\%$ of samples after 2 s to jointly satisfy planar-velocity error $\leq0.10$ m/s and payload tilt $\leq5^\circ$.
All methods use the same tolerances.
Planar RMSE averages episode-wise mean squared errors equally before taking the square root, including startup and terminal/failure samples.

\begin{table}[t]
\caption{B2--B2W best-checkpoint evaluation: seven nonzero commands, 448 episodes per seed. Mean $\pm$ sample SD across three training seeds; bold marks the best mean.}
\label{tab:main-results}
\centering
\footnotesize
\setlength{\tabcolsep}{1pt}
\begin{tabular*}{\columnwidth}{@{\extracolsep{\fill}}lrrr@{}}
\toprule
Method & \shortstack{Survival $\uparrow$\\(0--1)} & \shortstack{Tracking success $\uparrow$\\(0--1)} & \shortstack{$v$ RMSE $\downarrow$\\(m/s)}\\
\midrule
MAPPO & 0.966${}\pm{}$0.041 & 0.484${}\pm{}$0.049 & 0.0720${}\pm{}$0.0024\\
HAPPO & 0.867${}\pm{}$0.084 & 0.532${}\pm{}$0.059 & 0.1037${}\pm{}$0.0142\\
HASAC & 0.857${}\pm{}$0.146 & 0.510${}\pm{}$0.114 & 0.0777${}\pm{}$0.0111\\
HALO & 0.599${}\pm{}$0.167 & 0.066${}\pm{}$0.068 & 0.2188${}\pm{}$0.0101\\
\method{} (ours) & \textbf{1.000${}\pm{}$0.000} & \textbf{0.935${}\pm{}$0.074} & \textbf{0.0476${}\pm{}$0.0036}\\
\bottomrule
\end{tabular*}
\end{table}

On these seven commands, \method{} attains the highest mean velocity-tracking success rate $0.935\pm0.074$, the highest mean survival, and the lowest mean planar-velocity RMSE.

\subsection{Policy Representation Analysis}
\label{sec:policy-analysis}

To compare policy expressiveness, we construct a one-step bandit with action $a\in[-2,2]$ and reward
\begin{equation}
 r(a)=\sum_{\mu\in\{-1,1\}}
 \exp\!\left[-\frac{(a-\mu)^2}{2\sigma_r^2}\right],
 \qquad \sigma_r=0.3.
 \label{eq:bandit-reward}
\end{equation}
The reward has two separated high-value regions near $a=\pm1$.
At temperature $\alpha=0.1$, the corresponding Boltzmann optimum $\pi^\star(a)\propto\exp(r(a)/\alpha)$ is bimodal.
The diffusion actor used by \method{} and the squashed Gaussian actor used by HASAC are trained directly against the known reward, using their respective entropy regularizers.
This controlled setting removes critic-estimation error and partner-policy changes from the comparison.

Figure~\ref{fig:evidence}(a) shows the learned action distributions for three training seeds.
All three diffusion policies retain both modes; the Gaussian policies either spread probability across the low-reward interval or concentrate on one mode.
Using a common sample-based estimate of physical-action entropy for evaluation, the objective $\mathbb E[r]+\alpha\mathcal H$ is $0.83\pm0.00$ for the diffusion actor and $0.65\pm0.15$ for the Gaussian actor (mean $\pm$ standard deviation across seeds).

At one visited transport state (Fig.~\ref{fig:evidence}(b)), we sweep the robots' forward-velocity commands, fixing all remaining action coordinates, and evaluate the learned clipped double-Q function.
Two interior maxima correspond to both robots moving forward and to one robot reversing while the other advances, associated with translation and pivoting, respectively.
The value decreases along the straight segment between these maxima.
This task-specific value structure motivates a policy representation that can retain separated high-value responses; closed-loop transport performance is assessed in Section~\ref{sec:transport-performance}.

\subsection{Real-World Experiments}
\label{sec:hardware}

\subsubsection{Heterogeneous Robot Transport}
Figure~\ref{fig:transport-gallery}(g)--(k) shows a B2--B2W team transporting an unfixed rod during straight travel and turning.
Policies trained without an additional suspended load also support loaded transport without further training, demonstrating transfer to an unseen payload configuration.

\begin{figure}[t!]
  \centering
  \includegraphics[width=\columnwidth,trim=0 3bp 0 4bp,clip]{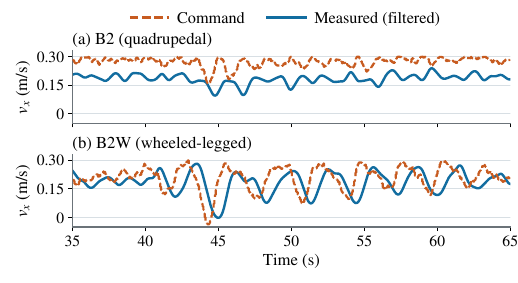}
  \caption{Real-world heterogeneous transport: high-level velocity commands and filtered odometry estimates for (a) quadrupedal B2 and (b) wheeled-legged B2W. B2 exhibits a persistent command--velocity offset, whereas B2W follows its varying commands more closely. Longitudinal velocities are expressed in each robot's body frame.}
  \label{fig:hardware-velocity}
\end{figure}

Figure~\ref{fig:hardware-velocity} shows a representative 30-s segment of straight transport.
B2's measured velocity generally remains below its command, whereas B2W follows its varying commands more closely.
This suggests that the high-level policies accommodate different low-level responses by issuing higher commands to B2 to compensate for its velocity shortfall, while modulating B2W's commands to maintain coordinated transport.

\subsubsection{Human--Robot Cooperative Transport}
Multi-agent training offers a route to learning robot policies for human--robot cooperation~\cite{zhang2026halypo}.
As shown in Fig.~\ref{fig:transport-gallery}(l), we examine this setting by pairing a B2W, controlled by a coordination policy trained with a robot teammate, with a human partner.
The human shares payload support and influences its motion through physical interaction.
A model predictive path integral controller converts a reference trajectory for the rod center into the team command $\mathbf c$ supplied to the high-level coordination policy.
Using this command and local robot and payload observations, the robot adjusts its motion to coordinate with the human through the shared payload.

\section{Conclusion}
\label{sec:conclusion}

This work introduced \method{}, a hierarchical multi-agent reinforcement learning framework for cooperative transport with heterogeneous robots supporting an unfixed shared payload. Robot-specific diffusion policies are coupled during training through a centralized team-value function, while decentralized execution relies only on local observations. A variational diffusion objective enables value-guided policy optimization without explicit action-density evaluation, and sequential updates account for evolving teammate policies.

The evaluations show that \method{} can represent distinct high-value action responses more effectively than Gaussian policies and improve cooperative transport performance. In simulation, it achieves the highest mean survival and velocity-tracking success among the evaluated baselines, with a tracking success rate of $0.935\pm0.074$ and a planar-velocity RMSE of $0.0476\pm0.0036$~m/s. The learned policies are also deployed on a heterogeneous quadrupedal--wheeled-legged team and with a human partner using only local robot and payload observations.

These results support the use of team-value-guided, robot-specific diffusion policies for coordinating heterogeneous embodiments. Future work will focus on larger teams, broader payload and environment variations, and more extensive quantitative real-world validation.

% \section{Conclusion}
% \label{sec:conclusion}

% \method{} learns coordination among robots with different motion capabilities for transport of an unfixed shared payload, without demonstrations or imitation pretraining.
% Robot-specific diffusion policies are trained from scratch through team-value guidance and variational optimization.
% The hierarchy separates coordination from robust command tracking, while onboard payload perception enables execution from local observations without partner-state inputs or inter-robot messages.
% Simulation comparisons show that \method{} achieves higher final mean returns and velocity-tracking success rates, as well as lower velocity errors, than the evaluated baselines.
% Hardware demonstrations further establish deployment with quadrupedal--wheeled-legged teams and human partners.

\appendices
\section{Theoretical Analysis}
\label{sec:analysis}

We first explain what the entropy replacement changes in the actor objective, then establish sequential improvement of the same surrogate under ideal policy iteration.
Temperatures and reference paths are fixed throughout.

\subsection{Entropy gap and response fitting}
\label{sec:entropy-gap}

\textbf{Proposition 1 (entropy gap).}
For compatible path densities with finite entropies, log-Jacobian expectations, and KL terms, let $p_i^0$ be the endpoint marginal and $p_i^c=p_i(z_i^{1:N}\mid z_i^0,o_i)$ the conditional path density.
Then
\begin{equation*}
 \mathcal H(\pi_i(\cdot\mid o_i))-\mathcal L_i(o_i)
 =\mathbb E_{z_i^0\sim p_i^0}D_{\mathrm{KL}}(p_i^c\|q_i^c)\geq0.
 \label{eq:elbo-identity-icra}
\end{equation*}
\emph{Proof sketch.}
Factor $p_i=p_i^0p_i^c$ in~\eqref{eq:path-score}.
The conditional log-density ratio contributes the negative expected KL divergence above, while change of variables gives
$-\mathbb E\log p_i^0+\mathbb E\log J_i=\mathcal H(\pi_i)$.
Their sum proves the identity.\hfill$\square$

To connect this result to the actor loss, fix $Q_\phi$ and teammate policies $\mu_{-i}$.
Substituting this identity and
$\log\mathcal B_i^\mu=U_i^\mu/\alpha_i-\log Z_i^\mu$ into~\eqref{eq:actor-loss} gives
\begin{equation*}
 \begin{split}
 \mathcal L_{\mathrm{actor}}^i
 &=\alpha_i\mathbb E_{o_i\sim\mathcal D}\!\Big[
 D_{\mathrm{KL}}(\pi_i\|\mathcal B_i^\mu)\\
 &\quad+\mathbb E_{z_i^0\sim p_i^0}D_{\mathrm{KL}}(p_i^c\|q_i^c)
 -\log Z_i^\mu(o_i)\Big],
 \end{split}
 \label{eq:actor-response-gap}
\end{equation*}
where both action distributions are conditioned on $o_i$.
The last term is constant during this coordinate update.
Training therefore balances response matching and the conditional-path gap; a finite sampler need not exactly realize the Boltzmann response.

\subsection{Monotonic improvement of the surrogate}

The following result adapts sequential soft policy improvement~\cite{liu2024maximum} to our path-based regularizer.
In this analysis, complete diffusion-path policies and value functions are conditioned on the Markov state $s$ rather than observation histories.

\textbf{Proposition 2 (sequential surrogate improvement).}
Assume bounded rewards, $0<\gamma<1$, fixed positive temperatures and reference paths, and uniformly bounded expected absolute path scores.
Let $Q=\proxyQ_\pi$ be the exact surrogate value of $\pi$ and define
$F_Q(\mu;s)=\mathbb E_{\mathbf a\sim\mu}Q(s,\mathbf a)+\sum_i\alpha_i\mathcal L_i(\mu_i;s)$,
where $\mathcal L_i(\mu_i;s)$ is the expected path score of policy $\mu_i$.
Starting from $\mu^{(0)}=\pi$, suppose each coordinate update holds the other current policies fixed and makes $F_Q$ nondecreasing at every state.
The completed sweep $\pi'=\mu^{(n)}$ then satisfies
\begin{equation*}
 \proxyQ_{\pi'}\geq\proxyQ_\pi,\qquad
 \proxyJ_\rho(\pi')\geq\proxyJ_\rho(\pi).
 \label{eq:improvement-icra}
\end{equation*}
\emph{Proof sketch.}
Chaining the coordinate inequalities gives
$F_Q(\pi';s)\geq F_Q(\pi;s)=\proxyV_\pi(s)$, the surrogate state value.
For a fixed path policy $\nu$, the surrogate Bellman operator
$\overline{\mathcal T}^{\nu}Q=r+\gamma\mathbb E_{s'\sim P(\cdot\mid s,\mathbf a)}F_Q(\nu;s')$ is monotone and a $\gamma$-contraction, since the regularizer cancels when comparing value functions.
Hence $\overline{\mathcal T}^{\pi'}\proxyQ_\pi\geq\proxyQ_\pi$, and repeated application converges monotonically to $\proxyQ_{\pi'}$.
Substitution into the state value and averaging over $\rho$ establish the return inequality.\hfill$\square$

% Balance the reference columns after final pagination.
\balance
\bibliographystyle{IEEEtran}
\bibliography{references}

\end{document}